\documentclass{article}
\pdfoutput=1

\usepackage{arxiv}

\usepackage[utf8]{inputenc} 
\usepackage[T1]{fontenc}    
\usepackage{hyperref}       
\usepackage{url}            
\usepackage{booktabs}       
\usepackage{amsfonts}       
\usepackage{nicefrac}       
\usepackage{microtype}      

\usepackage{xcolor}         
\usepackage{graphicx}       
\usepackage{subfigure}		
\usepackage{makecell}       
\usepackage{amsmath}        
\usepackage{xurl}           
\usepackage{lineno}  

\title{SAITS: Self-Attention-based Imputation for Time Series \thanks{Published in the journal \href{https://www.sciencedirect.com/journal/expert-systems-with-applications}{Expert Systems with Applications}. Its DOI link: \url{https://doi.org/10.1016/j.eswa.2023.119619}}}

\author{%
	Wenjie Du \thanks{Work was done during a machine-learning research internship at Ciena, co-funded by Mitacs.} \\
	Concordia University \\
	Montr\'eal, Canada \\
	\texttt{wenjay.du@gmail.com} \\
	\And
	David Cote \\
	Ciena Corporation \\
	Ottawa, Canada \\
	\texttt{dcote@ciena.com} \\
	\And
	Yan Liu \\
	Concordia University \\
	Montr\'eal, Canada \\
	\texttt{yan.liu@concordia.ca} \\
}

\begin{document}
\maketitle

\begin{abstract}
Missing data in time series is a pervasive problem that puts obstacles in the way of advanced analysis. A popular solution is imputation, where the fundamental challenge is to determine what values should be filled in. This paper proposes SAITS, a novel method based on the self-attention mechanism for missing value imputation in multivariate time series. Trained by a joint-optimization approach, SAITS learns missing values from a weighted combination of two diagonally-masked self-attention (DMSA) blocks. DMSA explicitly captures both the temporal dependencies and feature correlations between time steps, which improves imputation accuracy and training speed. Meanwhile, the weighted-combination design enables SAITS to dynamically assign weights to the learned representations from two DMSA blocks according to the attention map and the missingness information. Extensive experiments quantitatively and qualitatively demonstrate that SAITS outperforms the state-of-the-art methods on the time-series imputation task efficiently and reveal SAITS' potential to improve the learning performance of pattern recognition models on incomplete time-series data from the real world. The code is open source on GitHub at \url{https://github.com/WenjieDu/SAITS}.
\end{abstract}

\keywords{Time Series \and Missing Values \and Imputation Model \and Self-Attention \and Neural Network}

\section{Introduction} \label{intro}
Multivariate time-series data is ubiquitous in many application domains, for instance, transportation, economics, healthcare, and meteorology. State-of-the-art (SOTA) pattern recognition methods can well handle various time-series analysis tasks on complete datasets. However, due to all kinds of reasons, including failure of collection sensors, communication error, and unexpected malfunction, missing values are common to see in time series from the real-world environment, for example, patient health measurement~\cite{Silva2012ICU}, air-quality monitoring~\cite{Yi2016STMVL}, telecommunication networking~\cite{du2021ILOS}, and educational systems~\cite{BRUNI2021106512}. They impair the interpretability of data and pose challenges for advanced analysis and pattern recognition tasks, for instance, classification and clustering. To learn with such incomplete time series, thinking ahead of how to deal with missing parts before modeling is an inevitable step.

Traditional missing value processing methods fall into two categories. One is deletion, which removes samples or features that are partially observed. However, deletion makes data incomplete and can yield biased parameter estimates~\cite{Graham2009MissingData}. The other one is data imputation that estimates missing data from observed values~\cite{Rubin1975MissingData}. There are two main advantages to imputation over deletion~\cite{Burgess2013Imputation}: (1). Deletion introduces bias, while correctly specified imputation estimates are unbiased~\cite{White2010Imputation}; (2). Partially observed data may still be informative and helpful to the analysis. Nevertheless, the problem of imputation is what values should be filled in. Amounts of prior work are proposed to solve this problem with statistics and machine learning methods~\cite{Yu2016Temporal, Ansley1984ARIMA,Acuna2004Missing,Kreindler2006MissingData, Wang2006Unifying, Fung2006Methods, Azur2011ChainedEquations}. However, most of them require strong assumptions on missing data~\cite{Cao2018BRITS}, for example, linear regression~\cite{Ansley1984ARIMA}, mean/median averaging and k-nearest neighbors~\cite{Acuna2004Missing}. Such assumptions may introduce a strong bias that influences the outcome of the analysis. Recently, much literature utilizes deep learning to solve this imputation problem and achieves SOTA results. Non-time series imputation models, which are not in the scope of this paper, can be referred to~\cite{Yoon2018GAIN, Li2018MisGAN, Lee2019CollaGAN, Yoon2020GAMIN,Richardson2020McFlow,Nazabal2020HIVAE, Wu2020AimNet, Mouselinos2021MAIN}. A more detailed description of time-series imputation models is presented in Section~\ref{related_work}, including~\cite{Cao2018BRITS, Yoon2019MRNN, Liu2019NAOMI, Luo2018GRUI, Luo2019E2GAN, Fortuin2020GPVAE, Ramchandran2021LVAE, Ashman2020SGPVAE, Ma2019CDSA, Bansal2021DeepMVI, Shan2021NRTSI}.

Missing values can be characterized into three types: (1). Missing completely at random (MCAR), missing values are independent of any other values; (2). Missing at random (MAR), missing values depend only on observed values; (3). Missing not at random (MNAR), missing values depend on both observed and unobserved values~\cite{Rubin1976Missing,Little1986MissingData}. This work focuses on the MCAR case, as is standard in most of literature in the field of imputation. We uniformly sample values to be missing independently and introduce them as artificial missingness to evaluate all imputation methods used in this paper.

The self-attention mechanism is now widely applied, whereas its application on time-series imputation is still limited. Previous SOTA time-series imputation models are mostly based on recurrent neural networks (RNN), such as~\cite{Cao2018BRITS, Yoon2019MRNN, Luo2018GRUI, Luo2019E2GAN, Liu2019NAOMI}. Among them, the methods~\cite{Cao2018BRITS, Yoon2019MRNN, Luo2018GRUI, Luo2019E2GAN} are autoregressive models that are highly susceptible to compounding errors~\cite{Venkatraman2015TimeSeries,Liu2019NAOMI}. Although the work~\cite{Liu2019NAOMI} is not autoregressive, the multi-resolution imputation algorithm it proposed is made up of a loop, which can greatly slow the imputation speed. The self-attention mechanism, which is non-autoregressive and can overcome RNNs' drawbacks of slow speed and memory constraints, can avoid compounding error and be helpful to achieve better imputation quality and faster speed. This paper proposes a novel model called SAITS (\textit{\textbf{S}elf-\textbf{A}ttention-based \textbf{I}mputation for \textbf{T}ime \textbf{S}eries}) to learn missing values by a joint-optimization training approach of imputation and reconstruction. Particularly, our contributions in this work are summarized as the following:
\renewcommand{\theenumi}{\Roman{enumi}}
\begin{enumerate}
	\item We design a joint-optimization training approach of imputation and reconstruction for self-attention models to perform missing value imputation for multivariate time series. Transformer trained by this approach outperforms SOTA methods.
	\item We design a model called SAITS that consists of a weighted combination of two diagonally-masked self-attention (DMSA) blocks, which emancipates SAITS from RNN and enables it to capture temporal dependencies and feature correlations between time steps explicitly.
	\item We conduct adequate experiments and ablation studies on four real-world public datasets to quantitatively and qualitatively evaluate our methodology and justify its design. The experimental results do not only prove that SAITS achieves the new SOTA position for the imputation accuracy, but also show SAITS' potential of facilitating pattern recognition models to learn with partially-observed time series from the real world.
\end{enumerate}

The rest of this work is organized as follows: Section~\ref{related_work} reviews the related literature.  Section~\ref{methodology} introduces our joint-optimization training approach and the SAITS model. Experiments and conclusions are presented in Section~\ref{experiments} and~\ref{conclusion}, respectively.

\section{Related Work} \label{related_work}
We review the prior related work of time-series imputation in the following four categories:

\textbf{RNN-based} \hspace{1em} 
Che et al.~\cite{Che2018GRUD} propose GRU-D, a gated recurrent unit (GRU) variant, to handle missing data in time series classification problems. The concept of time decay on the last observation is firstly proposed by~\cite{Che2018GRUD} and continues to be used in~\cite{Cao2018BRITS, Yoon2019MRNN, Luo2018GRUI, Luo2019E2GAN}. M-RNN~\cite{Yoon2019MRNN} and BRITS~\cite{Cao2018BRITS} impute missing values according to hidden states from bidirectional RNN. However, M-RNN treats missing values as constants, while BRITS treats missing values as variables of the RNN graph. Furthermore, BRITS takes correlations among features into consideration while M-RNN does not.

\textbf{GAN-based} \hspace{1em}
Models in ~\cite{Luo2018GRUI,Luo2019E2GAN,Liu2019NAOMI} are also RNN-based. However, due to they adopt the generative adversarial network (GAN) structure, they are listed separately as GAN-based. Luo et al.~\cite{Luo2018GRUI} propose GRUI (GRU for Imputation) to model temporal information of incomplete time series. Both the generator and the discriminator in their GAN model are based on GRUI. Moreover, based on~\cite{Luo2018GRUI}, Luo et al.~\cite{Luo2019E2GAN} propose E$^2$GAN, which is an end-to-end method, comparing to the method in~\cite{Luo2018GRUI} having two stages. E$^2$GAN adopts an auto-encoder based on GRUI to form its generator to ease the difficulty of model training and improve imputation performance. Liu et al.~\cite{Liu2019NAOMI} propose a non-autoregressive model called NAOMI for spatiotemporal sequence imputation, which consists of a bidirectional encoder and a multiresolution decoder. NAOMI is further enhanced by adversarial training.

All the above RNN-based imputation models are susceptible to the recurrent network structure. They are time-consuming and have memory constraints, which make it hard for them to handle long-term dependency in time series, especially when the number of time steps in data samples is big. Apart from these disadvantages, most of these models are autoregressive that leads to the problem which is called compounding error in the field of time-series analysis~\cite{Venkatraman2015TimeSeries}. Although NAOMI alleviates this problem, its imputation algorithm consists of a loop, which can greatly slow the imputation speed together with its RNN structure. In addition, RNN models are usually unidirectional~\cite{Che2018GRUD,Luo2018GRUI,Luo2019E2GAN}. Even the models in~\cite{Cao2018BRITS,Yoon2019MRNN,Liu2019NAOMI} leverage bidirectional RNN to further improve the performance. Their final imputation results come from the average of two RNN models working on two directions (forward and backward) separately. As a result, these models are not deeply bidirectional~\cite{Devlin2019BERT}.

\textbf{VAE-based} \hspace{1em}
Inspired by GPPVAE~\cite{Casale2018GPPVAE} and the non-time-series imputation model HI-VAE~\cite{Nazabal2020HIVAE}, Fortuin et al.~\cite{Fortuin2020GPVAE} propose GP-VAE, a variational auto-encoder (VAE) architecture for time series imputation with a Gaussian process (GP) prior in the latent space. The GP-prior is used to help embed the data into a smoother and more explainable representation. L-VAE~\cite{Ramchandran2021LVAE} uses an additive multi-output GP-prior to accommodate auxiliary covariate information other than time. To support sparse GP approximations based on inducing points and handle missing values in spatiotemporal datasets, Ashman et al.~\cite{Ashman2020SGPVAE} propose SGP-VAE.

The GAN-based and VAE-based models are all generative ones, and they are difficult to train~\cite{Wu2020AimNet}. Particularly, GAN models suffer from the problems of non-convergence and mode collapse due to their loss formulation~\cite{Salimans2016GAN}. VAE models tend to involve latent variables used in the sampling and imputation, while they often do not correspond to concrete structures or distributions of the data, and this can make it difficult to interpret the imputation process for further understanding~\cite{Wu2020AimNet}.

\textbf{Self-Attention-based} \hspace{1em}
Ma et al.~\cite{Ma2019CDSA} apply cross-dimensional self-attention (CDSA) jointly from three dimensions (time, location, and measurement) to impute missing values in geo-tagged data, namely spatiotemporal datasets. Bansal et al.~\cite{Bansal2021DeepMVI} propose DeepMVI for missing value imputation in multidimensional time-series data. Their model includes a Transformer with a convolutional window feature and a kernel regression. Shan et al.~\cite{Shan2021NRTSI} propose NRTSI, a time-series imputation approach treating time series as a set of $(time, data)$ tuples. Such a design makes NRTSI applicable to irregularly-sampled time series. The method directly uses a Transformer encoder for modeling and achieves SOTA performance in their work. Such above prior literature explores applying self-attention in the field of time-series imputation. However, CDSA~\cite{Ma2019CDSA} is specifically designed for spatiotemporal data rather than general time series, and both CDSA and DeepMVI~\cite{Bansal2021DeepMVI} are not open-source, which makes it hard for other researchers to reproduce their methods and results. Regarding NRTSI~\cite{Shan2021NRTSI}, its algorithm design consists of two nested loops, which weaken the advantage of self-attention that is parallelly computational. Even worse, such loops can lead NRTSI to slow processing. Related work of self-attention-based models for time-series imputation is not much. There are some non-time series imputation models based on self-attention, such as AimNet~\cite{Wu2020AimNet} and MAIN~\cite{Mouselinos2021MAIN}.

\section{Methodology} \label{methodology}
Our methodology is made up of two parts: (1). the joint-optimization training approach of imputation and reconstruction; (2). the SAITS model, a weighted combination of two DMSA blocks.

\subsection{Joint-optimization Training Approach} \label{methodology: joint-optimization approach}
\begin{figure}[!htb]
	\centering
	\includegraphics[width=17cm]{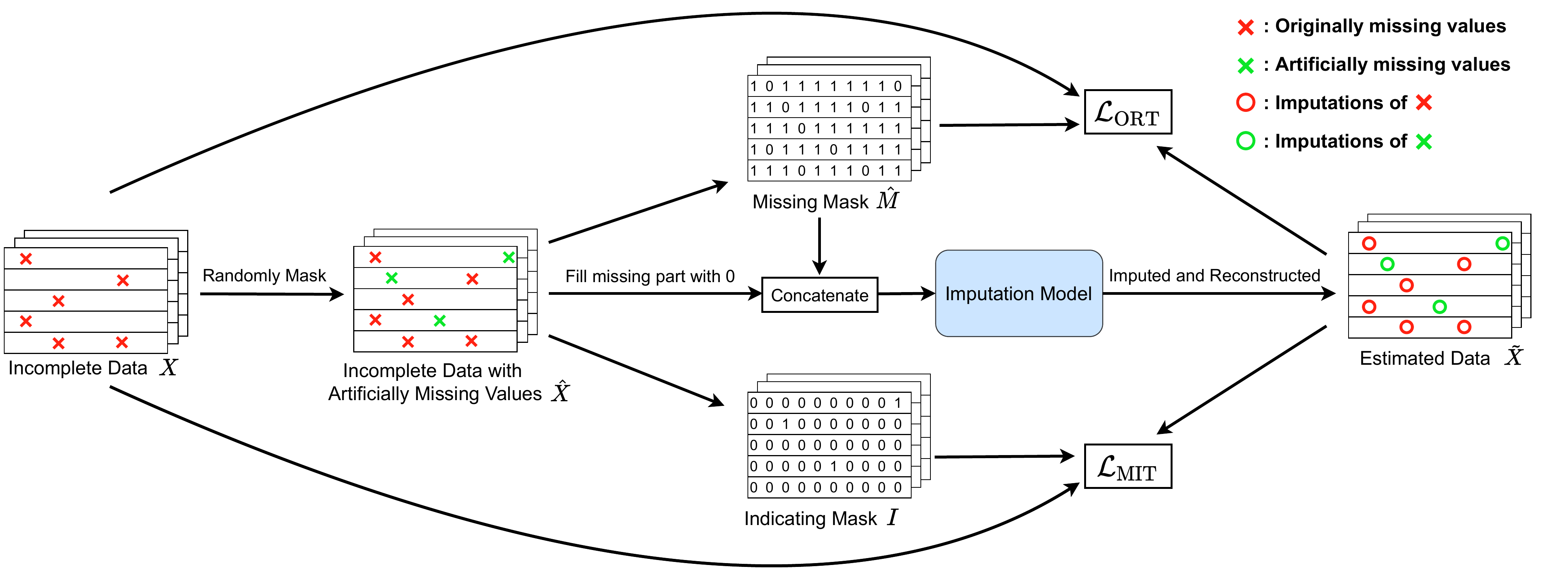}
	\caption{A graphical overview of the joint-optimization training approach, illustrating the implementation details of the two training tasks: masked imputation task (MIT) and observed reconstruction task (ORT).}
	\label{fig0}
\end{figure}

A general illustration of the joint-optimization approach is shown in Figure~\ref{fig0}. We are going to first give the definition of multivariate time series bearing missing data, then introduce the two learning objectives in detail.

\subsubsection{Definition of Multivariate Time-Series with Missing Values}
Given a collection of multivariate time series with $T$ time steps and $D$ dimensions, it is denoted as $X = \{x_1, x_2, ..., x_t, ..., x_T\} \in \mathbb{R}^{T \times D}$, where the $t$-th step $x_t = \{x_t^1, x_t^2, ..., x_t^d, ..., x_t^D\} \in \mathbb{R}^{1 \times D}$ and each value in it could be missing. Accordingly, $X_t^d$ represents the $d$-th dimension variable of the $t$-th step in $X$. To represent the missing variables in $X$, the missing mask vector $M \in \mathbb{R}^{T \times D}$ is introduced, where 

\begin{equation*}
	M_t^d = \left\{\begin{array}{ll}
		1 & \text{if } X_t^d \text{ is observed} \\
		0 & \text{if } X_t^d \text{ is missing} 
	\end{array}\right.
\end{equation*}

\subsubsection{Two Learning Tasks}
\begin{figure}[!b]
	\centering
	\subfigure[Imputation MAE]{ \label{fig1:a}
		\begin{minipage}[!t]{0.48\textwidth}
			\centering
			\includegraphics[width=8cm]{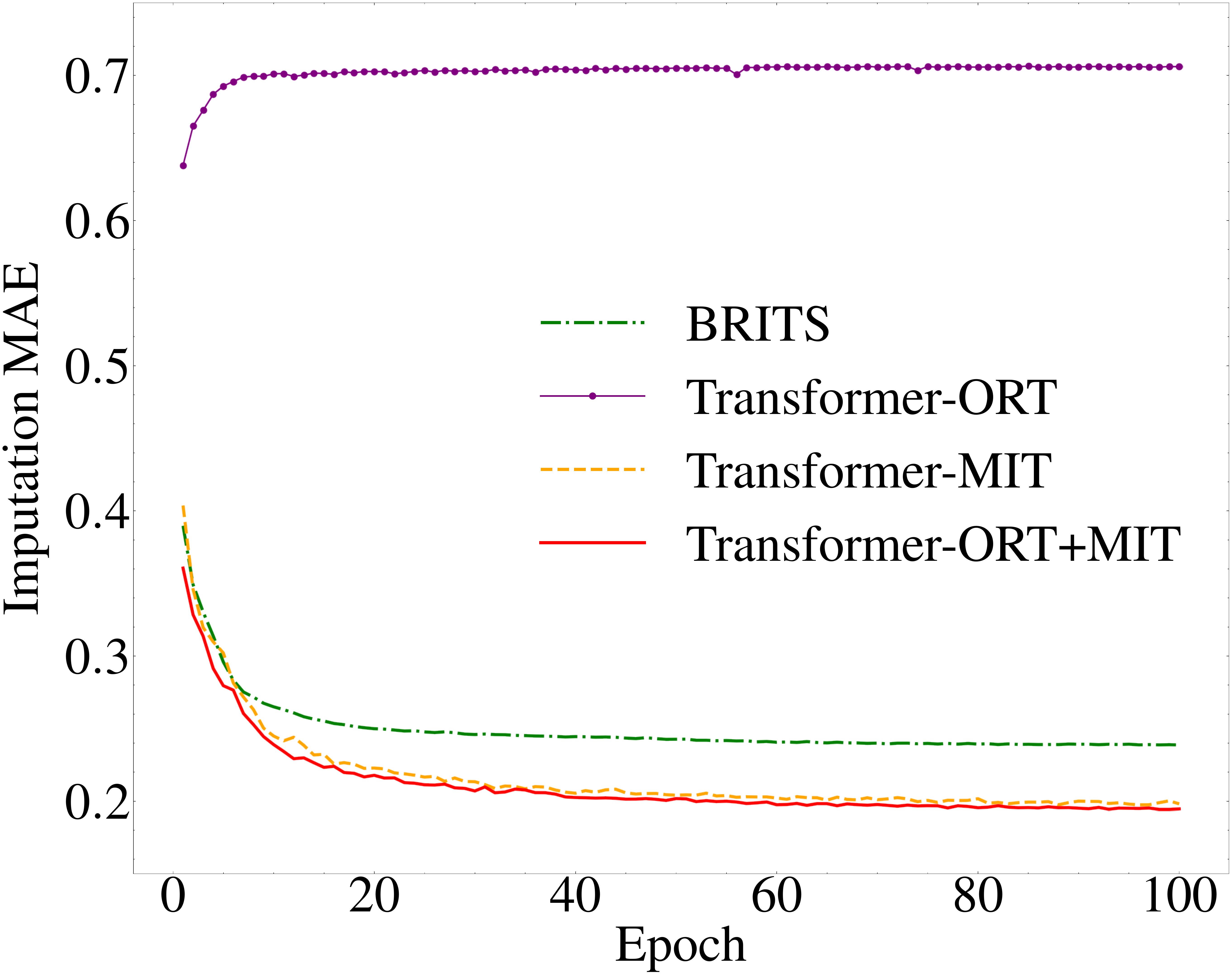}
		\end{minipage}
	}
	\subfigure[Reconstruction MAE]{ \label{fig1:b}
		\begin{minipage}[!t]{0.48\textwidth}
			\centering
			\includegraphics[width=8cm]{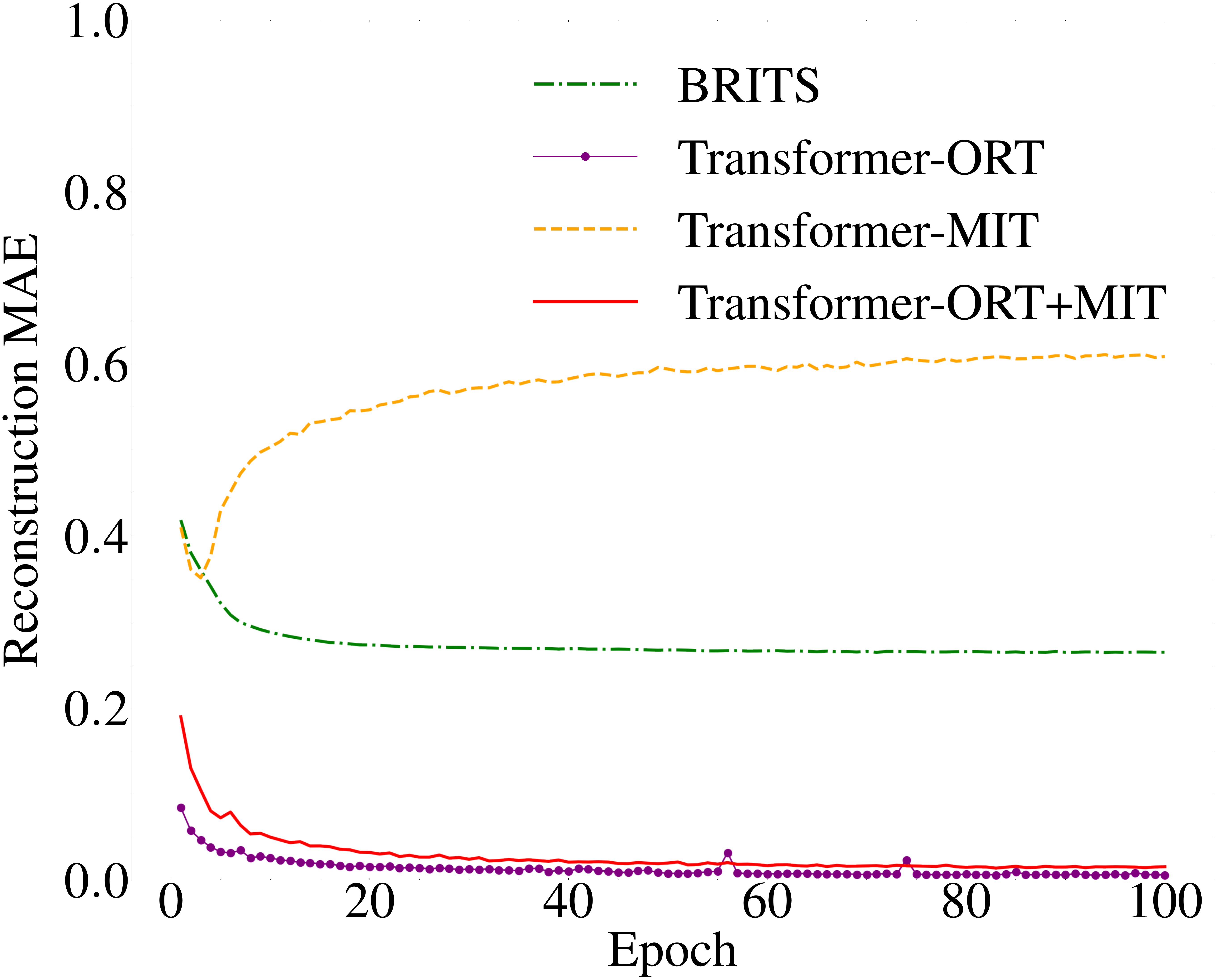}
		\end{minipage}
	}
	\caption{Imputation MAE and reconstruction MAE of the models in the validation stage. All models are trained on the same data. BRITS is trained with ORT, namely in the same way as its original paper. Transformer-ORT is trained with only ORT as well, i.e. without MIT. Transformer-MIT is trained with only MIT. Transformer-ORT+MIT is trained with the joint-optimization approach, namely with both ORT and MIT.}
	\label{fig1}
\end{figure}

To well train self-attention-based imputation models on the above defined multivariate time series with missing values, a joint-optimization training approach of imputation and reconstruction is designed. Here "imputation" is defined as the process in which the model fills the missing part in the given samples from null, and "reconstruction" means the model restores the observed values as exactly as possible after its processing. This joint-optimization approach consists of two learning tasks: Masked Imputation Task (MIT) and Observed Reconstruction Task (ORT). Correspondingly, the training loss is accumulated from two losses: the imputation loss of MIT and the reconstruction loss of ORT.

Before illustrating the joint-optimization approach in detail, it is necessary to discuss why we need a new training approach for self-attention-based imputation models. Here, for the straightforward visualization, BRITS~\cite{Cao2018BRITS} and Transformer~\cite{Vaswani2017SelfAttention} are taken as examples for explanation. The former stands for mainstream RNN-based imputation methods. The latter is a standard self-attention-based model. Figure~\ref{fig1} is plotted to make comparisons and to further illustrate the effects. 

A normal way to train an RNN for imputation consists of three main steps: (1). Input time-series feature vectors $X$ together with missing mask $M$ to alert the model that input data has observations and missing values; (2). Let the model reconstruct the observed part of the input time series and calculate the reconstruction error in each time step as the loss; (3). Finally, utilize the reconstruction loss to update the model. This training method is ORT. ORT works well with RNN-based models, for instance, BRITS. However, different from RNN which is autoregressive, self-attention itself is non-autoregressive and processes all input data parallelly and globally. Thus, if only trained on ORT, Transformer can distinguish the observed part from $X$ according to $M$, and as a result, it will focus only on minimizing the reconstruction error on the observed values. Taking a look at Figure~\ref{fig1:b}, Transformer-ORT's reconstruction MAE is much smaller than BRITS'. However, in Figure~\ref{fig1:a}, the imputation MAE of Transformer-ORT goes up from the beginning and is much larger than BRITS'. Transformer-ORT ignores the missing values because there is no penalty posed to it no matter what values it fills in. Hence, ORT can only ensure that Transformer gets well trained on observed values. In other words, there is no guarantee that Transformer-ORT can predict missing values accurately. To solve this optimization problem, we make another learning task MIT to become this guarantee and bind it together with ORT. This is how the joint-optimization training approach comes. The details of the two tasks are described as follows, and their concrete implementation is illustrated in Figure~\ref{fig1}.

\textbf{Task \#1: Masked Imputation Task (MIT)} \hspace{1em}
MIT is a prediction task on artificially-masked values, which explicitly forces the model to predict missing values accurately. In MIT, for every batch input into the model, some percentage (such as 20\% in our work) of observed values gets artificially masked at random. These values are not visible to the model, namely missing to the model. After artificially masking, the actual input time series is denoted as $\hat{X}$, and its corresponding missing mask vector is $\hat{M}$. The output estimated time series bearing reconstructions and imputations is denoted as $\tilde{X}$. To distinguish artificially-missing values and originally-missing values, the indicating mask vector $I$ is introduced. Math definitions of $\hat{M}$ and $I$ are: 

\begin{equation*}
	\hat{M}_t^d = \left\{\begin{array}{ll}
		1 & \text{if } \hat{X}_t^d \text{ is observed} \\
		0 & \text{if } \hat{X}_t^d \text{ is missing}
	\end{array}\right.,
	\hspace{1em}
	I_t^d = \left\{\begin{array}{ll}
		1 & \text{if } \hat{X}_t^d \text{ is artificially masked} \\
		0 & \text{otherwise}
	\end{array}\right. 
\end{equation*}

After the model imputes all missing values, the imputation loss is the mean absolute error (MAE) calculated between the artificially missing values and their respective imputations. The calculation of MAE and MIT loss are defined in Equation~\ref{mae_loss} and Equation~\ref{imputation_loss_training} below.

\begin{equation}
	\label{mae_loss}
	\ell_{\text{MAE}}\left(estimation, target, mask \right) = \frac{\sum_{d=1}^D \sum_{t=1}^T \lvert \left(estimation - target \right)\odot mask\rvert_t^d}{\sum_{d=1}^D \sum_{t=1}^T mask_t^d}
\end{equation}
\begin{equation}
	\label{imputation_loss_training}
	\mathcal{L}_\text{MIT} = \ell_{\text{MAE}}\left(\tilde{X}, X, I\right) 
\end{equation}

Note that learning tasks similar to MIT, which mask some objects and then predict them, are commonly used to train models in NLP (Natural Language Processing) field, for example, the Cloze task~\cite{Taylor1953ClozeTask}, and the Masked Language Modeling (MLM) used to pre-train BERT~\cite{Devlin2019BERT}. MIT is inspired by MLM, but the differences are: (1). MLM predicts missing tokens (time steps), while MIT predicts missing values in time steps; (2). One disadvantage of MLM is that it causes pretrain-finetune discrepancy because masking symbols used during pretraining are absent from real data in finetuning~\cite{Yang2019XLNet}. However, the original objective of imputation is to predict missing or masked values. Therefore, MIT does not cause such discrepancies. 

\textbf{Task \#2: Observed Reconstruction Task (ORT)} \hspace{1em}
ORT is a reconstruction task on the observed values. It is widely applied in the training of imputation models for both time-series and non-time series ~\cite{Cao2018BRITS,Luo2018GRUI,Luo2019E2GAN,Fortuin2020GPVAE,Yoon2018GAIN,Li2018MisGAN}. After model processing, observed values in the output are different from their original values, and they are called reconstructions. In our work, the reconstruction loss is MAE calculated between the observed values and their respective reconstructions, defined in Equation~\ref{reconstruction_loss_training} as below.

\begin{equation}
	\label{reconstruction_loss_training}
	\mathcal{L}_\text{ORT} = \ell_{\text{MAE}}\left(\tilde{X}, X, \hat{M}\right)
\end{equation}

In our training approach, MIT and ORT are integral. MIT is utilized to force the model to predict missing values as accurately as possible, and ORT is leveraged to ensure that the model converges to the distribution of observed data. As shown in Figure~\ref{fig1}, both the imputation MAE and reconstruction MAE of Transformer-ORT+MIT drop steadily. A comparison between Transformer-MIT and Transformer-ORT+MIT tells that MIT makes the main contribution to decreasing the imputation MAE. Compared with Transformer-ORT+MIT, Transformer-MIT has a slightly higher imputation MAE. This proves that ORT can help models further optimize performance on the imputation task. On the reconstruction MAE, Transformer-MIT climes up because it is not required to converge on the observed data. Furthermore, in the aspect of the reconstruction MAE in Figure~\ref{fig1:b}, Transformer-ORT+MIT is slightly higher than Transformer-ORT because the gradient of the reconstruction loss gets influenced by the imputation loss. This is another piece of evidence proving that our joint-optimization approach works.

It is worth mentioning that our training approach is designed not only for time-series self-attention models but can be applied to train other imputation models. As shown in Figure~\ref{fig0}, the imputation model in the blue box is not specified and can be replaced with other models for training. Moreover, the data can be non-time-series. We also discuss applying the joint-optimization approach to train BRITS in Appendix~\ref{BRITS_apply_MIT}.

\subsection{SAITS} \label{methodology: SAITS}
\begin{figure}[!ht]
	\centering
	\includegraphics[width=16cm]{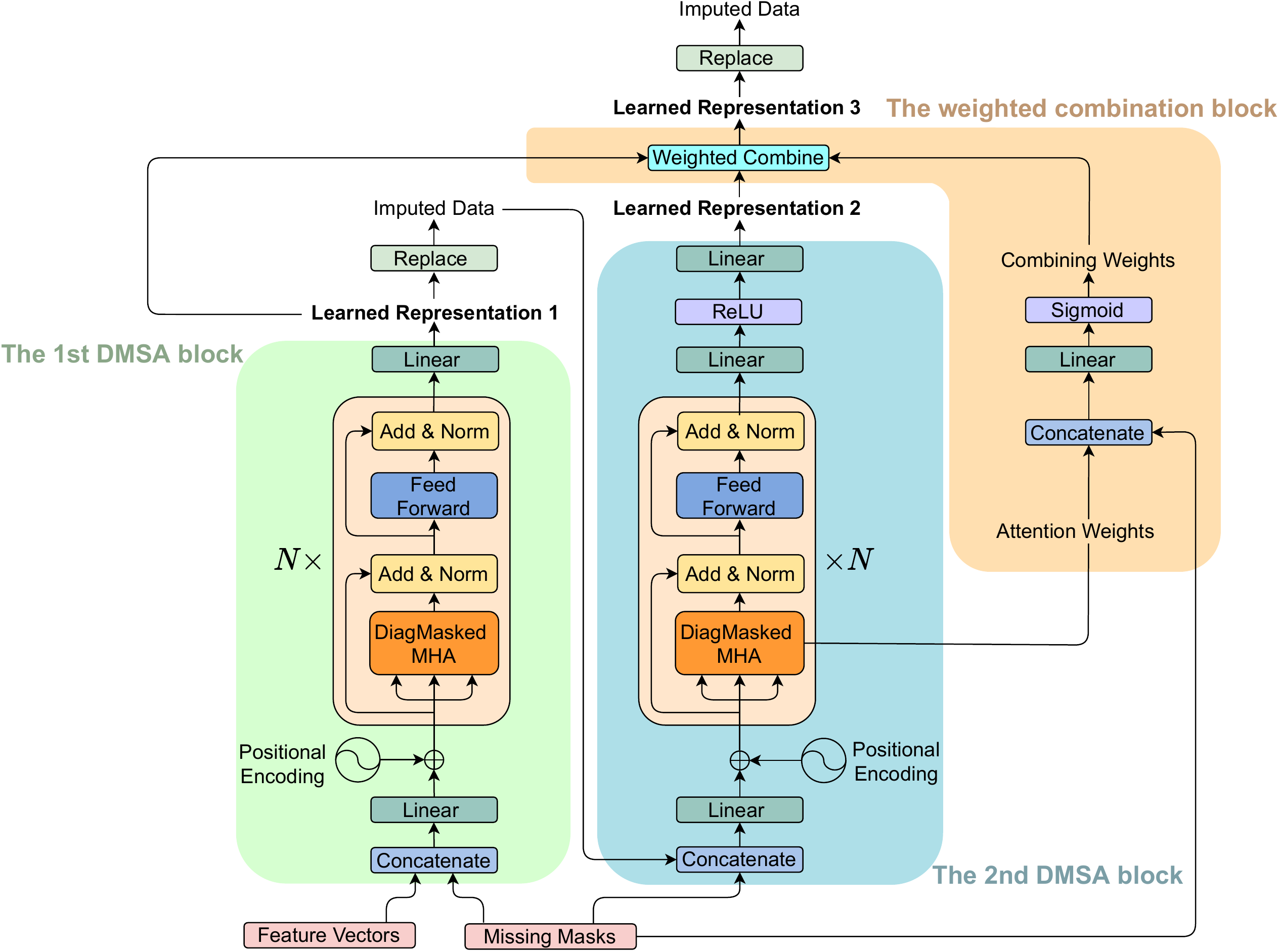}
	\caption{The SAITS model architecture.}
	\label{fig:SAITS_structure}
\end{figure}

As illustrated in Figure~\ref{fig:SAITS_structure}, SAITS is composed of two diagonally-masked self-attention (DMSA) blocks and a weighted combination. Firstly, some fundamental components of SAITS get introduced in Subsection~\ref{SAITS: diagonally-masked SA} and~\ref{SAITS: PE and FFN}. Then SAITS' three-part structure is illustrated in Subsection~\ref{SAITS: the 1st block},~\ref{SAITS: the 2nd block}, and~\ref{SAITS: the weighted combination block}, respectively. Finally, the loss functions of learning tasks are discussed in Subsection~\ref{SAITS: training loss}.

\subsubsection{Diagonally-Masked Self-Attention} \label{SAITS: diagonally-masked SA}
The conventional self-attention mechanism is proposed by Vaswani et al.~\cite{Vaswani2017SelfAttention} to solve the language translation task. Now it is widely applied in sequence modeling. A given sequence is mapped into a query vector $Q$ of dimension $d_k$, a key vector $K$ of dimension $d_k$ and a value vector $V$ of dimension $d_v$. The scaled dot-product can effectively calculate attention scores (or the attention map) between $Q$ and $K$. After that, a softmax function is applied to obtain attention weights. The final output is attention-weighted $V$. The whole process is as shown in Equation~\ref{SelfAttention} below:

\begin{equation}
	\label{SelfAttention}
	\operatorname{SelfAttention}\left(Q, K, V\right) =\operatorname{Softmax}\left(\frac{Q K^\mathsf{T}}{\sqrt{d_{k}}}\right) V
\end{equation}

\begin{figure}[!h]
	\centering
	\includegraphics[height=7.5cm]{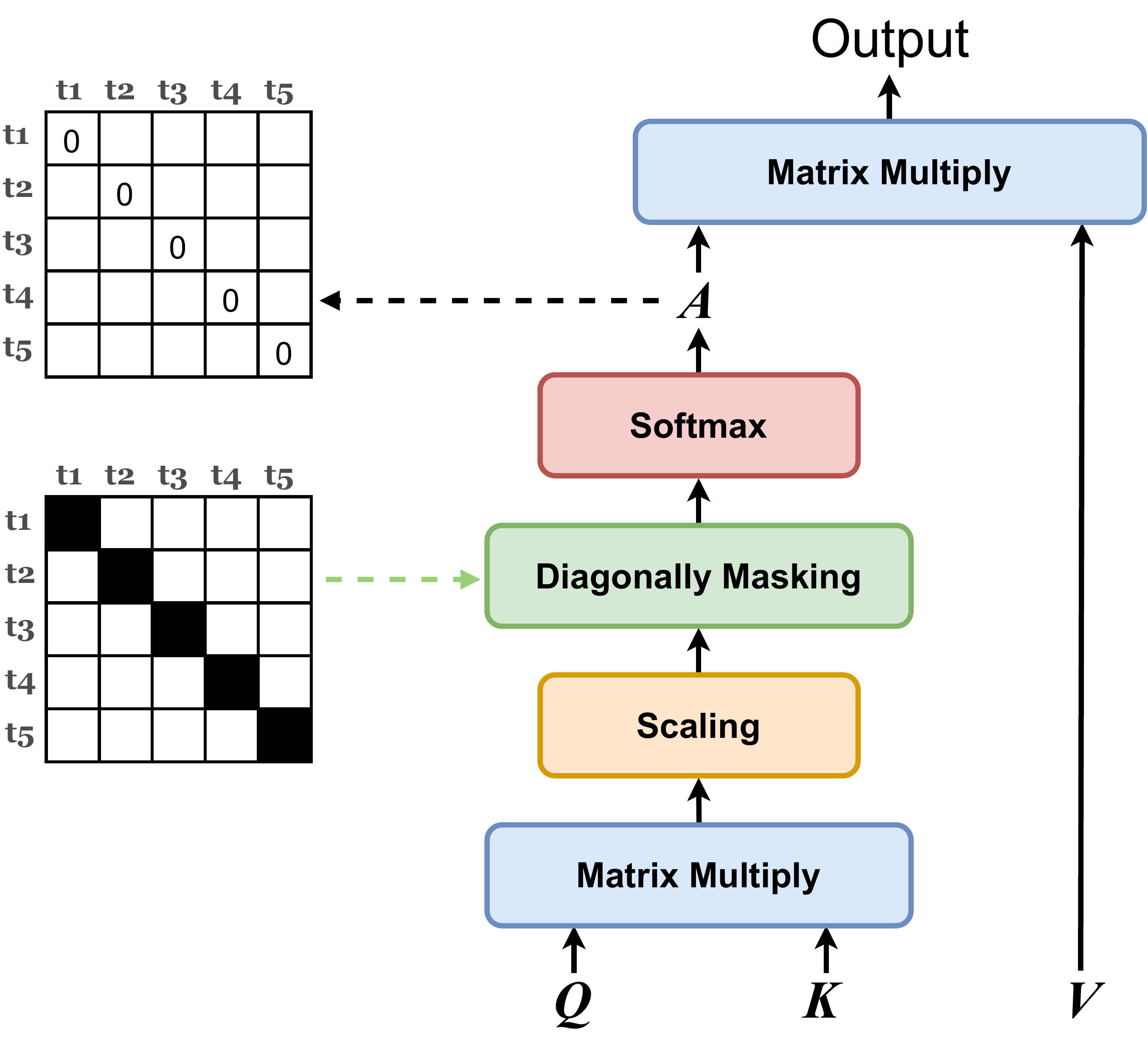}
	\caption{Illustration of Equation~\ref{DiagMaskedSelfAttention}. Diagonally-masked self-attention on a time-series sample with five time steps. }
	\label{fig:DiagMaskedSelfAttention}
\end{figure}

To enhance SAITS' imputation ability, the diagonal masks are applied inside the self-attention. As formulated in Equation~\ref{DiagMask} and~\ref{DiagMaskedSelfAttention}, the diagonal entries of the attention map ($\in \mathbb{R}^{T \times T}$) are set as $-\infty$ (set as $-1 \times 10^9$ in practice) so the diagonal attention weights approach 0 after the softmax function. Figure~\ref{fig:DiagMaskedSelfAttention} gives a vivid illustration of the DMSA mechanism.

\begin{equation} \label{DiagMask}
	\left[ \operatorname{DiagMask}\left(x\right) \right]\left(i, j\right) = \left\{\begin{array}{ll}
		-\infty & i=j \\
		x \left(i, j\right) & i\neq j
	\end{array}\right.
\end{equation}

\begin{equation} \label{DiagMaskedSelfAttention}
	\begin{aligned}
		\operatorname{DiagMaskedSelfAttention}\left(Q, K, V\right) &=\operatorname{Softmax}\left(\operatorname{DiagMask}\left(\frac{Q K^\mathsf{T}}{\sqrt{d_{k}}}\right)\right) V \\
		&=A V, \ \ \text{where } A \text{ is attention weights} \\
	\end{aligned}
\end{equation}

With these diagonal masks, input values at the $t$-th step can not see themselves and are prohibited from contributing to their own estimations. Consequently, their estimations depend only on the input values from other $(T-1)$ time steps. Such a mechanism makes DMSA able to capture the temporal dependencies and feature correlations between time steps in the high dimensional space with only one attention operation. Subsequently, the diagonally-masked multi-head attention ($\operatorname{DiagMaskedMHA}$) is formulated as:

\begin{equation} \label{DiagMaskedMHA}
	\begin{aligned}
		\operatorname{DiagMaskedMHA}\left(x\right) =\operatorname{Concat}&\left(head_1, head_2, ..., head_i, ..., head_h\right) W^O \\
		\text{where } head_i =\operatorname{DiagMaskedSelfAttention}&\left(x W_{i}^{Q}, x W_{i}^{K}, x W_{i}^{V}\right), h \text{ is the number of heads}
	\end{aligned}
\end{equation}

In the above Equation~\ref{DiagMaskedMHA}, $W_{i}^{Q} \in \mathbb{R}^{d_{\text {model }} \times d_{k}}$, $W_{i}^{K} \in \mathbb{R}^{d_{\text{model }} \times d_{k}}$, and $W_{i}^{V} \in \mathbb{R}^{d_{\text {model}} \times d_{v}}$ are parameters of the linear layers projecting input $x$ to $Q$, $K$, and $V$ separately. $W^{O} \in \mathbb{R}^{hd_v \times d_{\text {model}}}$ is the parameter of the output layer in $\operatorname{DiagMaskedMHA}$. 

To prove the effectiveness of DMSA, an ablation study is performed in Section~\ref{diagonal_mask_ablation_study}. Note that attention masks are widely applied in self-attention modeling, especially in NLP field, including the diagonal masks used here, for example~\cite{Shen2018DISAN, Yang2019XLNet, Shin2020FA}.

\subsubsection{Positional Encoding and Feed-Forward Network} \label{SAITS: PE and FFN}
In Transformer, Vaswani et al.~\cite{Vaswani2017SelfAttention} apply the positional encoding to make use of the sequence order because there is no notion of sequence order in the original Transformer architecture. Additionally, there is a fully-connected feed-forward network applied behind each attention layer. In SAITS, both the positional encoding and the feed-forward network are kept. 

The positional encoding consists of sine and cosine functions, which is formulated as Equation~\ref{positional_encoding} below. Note that $p$ is used to refer to the positional encoding in the following equations for brevity.

\begin{equation} \label{positional_encoding}
	\begin{aligned}
		\operatorname{PosEnc}(pos, 2i)   = \sin \left(\frac{pos}{10000^{\frac{2i}{d_{\text{model}}}}}\right),&\hspace{1em}
		\operatorname{PosEnc}(pos, 2i+1) = \cos \left(\frac{pos}{10000^{\frac{2i}{d_{\text{model}}}}}\right) \\
		\text{where } pos \text{ is the time-step } &\text{position, } i \text{ is the dimension}
	\end{aligned}
\end{equation}

The feed-forward network has two linear transformations with a ReLU activation function between them, as shown in Equation~\ref{feed_forward_network}:

\begin{equation} \label{feed_forward_network}
	\begin{aligned}
		\operatorname{FFN}\left(x\right) &= \operatorname{ReLU}\left(x W_1 + b_1\right) W_2 + b_2 \\
		\text{where } W_1 \in \mathbb{R}^{d_\text{model} \times d_\text{ffn}}&, b_1 \in \mathbb{R}^{d_\text{ffn}}, W_2 \in \mathbb{R}^{d_\text{ffn} \times d_\text{model}}, b_2 \in \mathbb{R}^{d_\text{model}}
	\end{aligned}
\end{equation}

\subsubsection{The First DMSA Block} \label{SAITS: the 1st block}
\begin{equation} \label{first_projection}
	e = \left[\operatorname{Concat}\left(\hat{X}, \hat{M}\right) W_e + b_e\right] + p
\end{equation}
\begin{equation} \label{first_block}
	z = \{\operatorname{FFN}(\operatorname{DiagMaskedMHA}\left(e\right))\}^N
\end{equation}
\begin{equation} \label{first_learned_representation}
	\tilde{X}_1 = z W_z + b_z 
\end{equation}
\begin{equation} \label{first_replace}
	\hat{X}' = \hat{M}\odot \hat{X} + \left(1 - \hat{M}\right)\odot \tilde{X}_1
\end{equation} 
In the first DMSA block, the actual input feature vector $\hat{X}$ and its missing mask vector $\hat{M}$ are concatenated as the input. Equation~\ref{first_projection} projects the input to $d_{\text{model}}$ dimensions and adds up with the positional encoding $p$ to produce $e$. $W_e$ and $b_e$ are parameters ($W_e \in \mathbb{R}^{2D \times d_{\text{model}}}$, $b_e \in \mathbb{R}^{d_{\text{model}}}$). Operation $\{\}^N$ in Equation~\ref{first_block} means stacking $N$ layers. Equation~\ref{first_block} transfers $e$ to $z$ with $N$ stacked layers of the diagonally-masked multi-head attention and the feed-forward network~\footnote{Note that the layer normalization~\cite{Ba2016LN} and residual connection~\cite{He2016ResNet} are applied after each attention layer and feed-forward layer in the same way as~\cite{Vaswani2017SelfAttention}. Figure~\ref{fig:SAITS_structure} shows these details. They are suppressed here for simplicity.}. Equation~\ref{first_learned_representation} reduces $z$ from $d_\text{model}$ dimensions to $D$ dimensions and produces $\tilde{X}_1$ (\textbf{Learned Representation 1}). Parameter $W_z \in \mathbb{R}^{d_{\text{model}} \times D}$ and $b_z \in \mathbb{R}^{D}$. In Equation~\ref{first_replace}, missing values in $\hat{X}$ are replaced with corresponding values in $\tilde{X}_1$ to obtain the completed feature vector $\hat{X}'$ with the observed part in $\hat{X}$ kept intact. Here, $\odot$ is the Hadamard product, also known as the element-wise product.

\subsubsection{The Second DMSA Block} \label{SAITS: the 2nd block}
\begin{equation}
	\label{second_projection}
	\alpha = \left[\operatorname{Concat}\left(\hat{X}', \hat{M}\right) W_{\alpha} + b_{\alpha}\right] + p
\end{equation} 
\begin{equation} 
	\label{second_block}
	\beta = \{\operatorname{FFN}\left(\operatorname{DiagMaskedMHA}\left(\alpha\right)\right)\}^N
\end{equation}
\begin{equation} 
	\label{second_learned_representation}
	\tilde{X}_2 = \operatorname{ReLU}\left(\beta W_{\beta} + b_{\beta}\right) W_{\gamma} + b_{\gamma}
\end{equation} 
The second DMSA block takes the output $\hat{X}'$ of the first DMSA block and continues learning. Similar to Equation~\ref{first_projection}, Equation~\ref{second_projection} projects the concatenation of $\hat{X}'$ and $\hat{M}$ from $D$ dimensions to $d_{\text{model}}$ dimensions and then adds the result together with $p$ to generate $\alpha$. Parameter $W_\alpha \in \mathbb{R}^{2D \times d_{\text{model}}}$, $b_\alpha \in \mathbb{R}^{d_{\text{model}}}$. Equation~\ref{second_block} performs $N$ times of nested attention functions and feed-forward networks on $\alpha$ and outputs $\beta$. In Equation~\ref{second_learned_representation}, to obtain $\tilde{X}_2$ (\textbf{Learned Representation 2}), two linear projections are applied on $\beta$ with a ReLU activation in between, where parameter $W_\beta \in \mathbb{R}^{d_{\text{model}} \times D}$, $b_\beta \in \mathbb{R}^{D}$, $W_\gamma \in \mathbb{R}^{D \times D}$, $b_\gamma \in \mathbb{R}^{D}$. Empirically, a deeper structure can learn better a representation to capture more complicated correlations in time series. Here, in Equation~\ref{second_learned_representation}, we apply one more non-linear layer than Equation~\ref{first_learned_representation} to build a deeper block. In practice, such an operation does help achieve a better imputation performance than applying a single linear projection. The same transformation is not applied to obtain $\tilde{X}_1$ in the first DMSA block because the learnable parameters in the following weighted combination can dynamically adjust the weights for $\tilde{X}_1$ and $\tilde{X}_2$ to form better $\tilde{X}_3$ (\textbf{Learned Representation 3}). Moreover, we find that even applying the same transformation here to obtain $\tilde{X}_1$ does not help achieve better results than the current design. It validates the effectiveness of our weighted combination, described as below.

\subsubsection{The Weighted Combination Block} \label{SAITS: the weighted combination block}
\begin{equation} \label{gene_attn_weights}
	\hat{A} = \frac{1}{h} \sum_{i=1}^h A_i
\end{equation}
\begin{equation} \label{combining_weights}
	\eta = \operatorname{Sigmoid}\left(\operatorname{Concat}\left(\hat{A}, \hat{M} \right) W_{\eta} + b_{\eta} \right)
\end{equation}
\begin{equation} \label{weighted_combine}
	\tilde{X}_3 = \left(1 - \eta \right) \odot \tilde{X}_1 + \eta \odot \tilde{X}_2
\end{equation}
\begin{equation} \label{second_replace}
	\hat{X}_c = \hat{M}\odot \hat{X} + \left(1 - \hat{M} \right)\odot \tilde{X}_3
\end{equation} 
To obtain a better learned representation $\tilde{X}_3$, the weighted combination block is designed to dynamically weigh $\tilde{X}_1$ and $\tilde{X}_2$ according to temporal dependencies and missingness information. $\hat{A}$ ($\in \mathbb{R}^{T \times T}$) in Equation~\ref{gene_attn_weights} is averaged from attention weights $A$ output by multi heads in the last layer of the second DMSA block. Equation~\ref{combining_weights} takes averaged attention weights $\hat{A}$ and missing masks $\hat{M}$ as references to produce the combining weights $\eta$ ($\in (0, 1)^{T \times D}$) with the learnable parameters $W_\eta$ ($\in \mathbb{R}^{(T+D) \times D}$) and $b_\eta$ ($\in \mathbb{R}^{D}$). Equation~\ref{weighted_combine} combines $\tilde{X}_1$ and $\tilde{X}_2$ by weights $\eta$ to form $\tilde{X}_3$. Finally, in Equation~\ref{second_replace}, missing values in $\hat{X}$ are replaced with corresponding values in $\tilde{X}_3$ to produce the complement vector $\hat{X}_c$, i.e. the imputed data. To further discuss the rationality of the weighted combination, an ablation experiment is performed in Section~\ref{weighted_combination_ablation_study}.

Moreover, the second DMSA block and the weighted combination block are added to extend the learning process of our model and to obtain better performance. We do not apply more than two DMSA blocks because the benefit brought is marginal. Experiments and analysis are conducted to prove our points here in Section~\ref{why_not_more_than_2_blocks}.

\subsubsection{Loss Functions of Learning Objectives} \label{SAITS: training loss}
\begin{equation}
	\label{reconstruction_loss}
	\mathcal{L}_\text{ORT} = \frac{1}{3} \left(
	\ell_{\text{MAE}}\left(\tilde{X}_1, X, \hat{M}\right)
	+ \ell_{\text{MAE}}\left(\tilde{X}_2, X, \hat{M}\right)
	+ \ell_{\text{MAE}}\left(\tilde{X}_3, X, \hat{M}\right) 
	\right)
\end{equation}
\begin{equation}
	\label{imputation_loss}
	\mathcal{L}_\text{MIT} = \ell_{\text{MAE}}\left(\hat{X}_c, X, I\right) 
\end{equation}
\begin{equation}
	\label{final_loss}
	\mathcal{L} = \mathcal{L}_\text{ORT} + \lambda \, \mathcal{L}_\text{MIT}
\end{equation}

There are two learning tasks in the model training: MIT and ORT. The imputation loss of MIT ($\mathcal{L}_\text{MIT}$) and the reconstruction loss of ORT ($\mathcal{L}_\text{ORT}$) are both calculated by the MAE loss function ($\ell_{\text{MAE}}$) defined in Equation~\ref{mae_loss}, which takes three inputs: $estimation$, $target$, and $mask$ (all of them $\in \mathbb{R}^{T \times D}$). It calculates MAE between values indicated by $mask$ in $estimation$ and $target$. $target$ and $mask$ of $\mathcal{L}_\text{ORT}$ in Equation~\ref{reconstruction_loss} are the input feature vector $\hat{X}$ and its missing mask vector $\hat{M}$. We let $\tilde{X}_1$ and $\tilde{X}_2$ directly participate in the composition of $\tilde{X}_3$. Accordingly, here $\mathcal{L}_\text{ORT}$ is accumulated from three learned representations: $\tilde{X}_1$, $\tilde{X}_2$ and $\tilde{X}_3$. Such an accumulated loss can lead to faster convergence speed. To ensure $\mathcal{L}_\text{ORT}$ not too large to dominate the direction of the gradient, it is reduced by a factor of three, i.e. averaged. Inputs $estimation$, $target$ and $mask$ of $\mathcal{L}_\text{MIT}$ in Equation~\ref{imputation_loss} are the complement feature vector $\hat{X}_c$, the original feature vector $X$ without artificially-masked values, and the indicating mask vector $I$, respectively. At last, Equation~\ref{final_loss} adds $\mathcal{L}_\text{ORT}$ and $\mathcal{L}_\text{MIT}$ together by a weighted sum, where $\lambda$ is the weighting coefficient that can be tuned. $\lambda$ is fixed as 1 in our experiments. Our SAITS model is updated by minimizing the final loss $\mathcal{L}$.

\section{Experiments}\label{experiments}
For the sake of the reproducibility of our results, we make our work publicly available to the community. Our data preprocessing scripts, model implementations, as well as hyper-parameter search configurations, are all available in the GitHub repository~\url{https://github.com/WenjieDu/SAITS}.

\subsection{Datasets}
In order to benchmark the proposed SAITS model, the experiments are performed on four public real-world datasets from different domains: PhysioNet 2012 Mortality Prediction Challenge~\footnote{\url{https://www.physionet.org/content/challenge-2012}}~\cite{Silva2012ICU}, Beijing Multi-Site Air-Quality~\footnote{\url{https://archive.ics.uci.edu/ml/datasets/Beijing+Multi-Site+Air-Quality+Data}}~\cite{Zhang2017AirQuality}, Electricity Load Diagrams~\footnote{\url{https://archive.ics.uci.edu/ml/datasets/ElectricityLoadDiagrams20112014}}~\cite{Dua2017UCI}, and Electricity Transformer Temperature~\footnote{\url{https://github.com/zhouhaoyi/ETDataset}}~\cite{Zhou2021informer}.

The descriptions of four datasets used in this work and their preprocessing details are elaborated as below. General information of all datasets is listed in Table~\ref{dataset_info}. Note that standardization is applied in the preprocessing of all datasets.
\begin{table} [!htb]
	\caption{General information of four datasets used in this work.}
	\label{dataset_info}
	\centering
	\begin{tabular}{p{100pt}<{\centering}|p{70pt}<{\centering}|p{70pt}<{\centering}|p{70pt}<{\centering}|p{70pt}<{\centering}}
		\toprule
		& PhysioNet-2012 & Air-Quality  & Electricity & ETT \\
		\midrule
		Number of total samples&  11,988        &  1,461       & 1,400         &  5,803          \\ 
		\midrule
		Number of features     &   37           &  132         & 370                     &       7  \\
		\midrule
		Sequence length        &   48           & \phantom{0}24& 100           &  24  \\ 
		\midrule
		Original missing rate  &  80.0\%        &  1.6\%       & 0\%               &     0\%  \\ 
		\bottomrule
	\end{tabular}
\end{table}

\textbf{PhysioNet 2012 Mortality Prediction Challenge (PhysioNet-2012)} \hspace{1em} 
The PhysioNet 2012 challenge dataset~\cite{Goldberger2000PhysioNet} contains 12,000 multivariate clinical time-series samples collected from patients in ICU (Intensive Care Unit). Each sample is recorded during the first 48 hours after admission to the ICU. Depending on the status of patients, there are up to 37 time-series variables measured, for instance, temperature, heart rate, blood pressure. Measurements might be collected at regular intervals (hourly or daily), and also may be recorded at irregular intervals (only collected as required). Not all variables are available in all samples. Note that this dataset is very sparse and has 80\% missing values in total. The dataset is firstly split into the training set and the test set according to 80\% and 20\%. Then 20\% of samples are split from the training set as the validation set. We randomly eliminate 10\% of observed values in the validation set and the test set and use these values as ground truth to evaluate the imputation performance of models. Following 12 samples are dropped because of containing no time-series information at all: 147514, 142731, 145611, 140501, 155655, 143656, 156254, 150309, 140936, 141264, 150649, 142998.

\textbf{Beijing Multi-Site Air-Quality (Air-Quality)}\hspace{1em} 
This air-quality dataset~\cite{Zhang2017AirQuality} includes hourly air pollutants data from 12 monitoring sites in Beijing. Data is collected from 2013/03/01 to 2017/02/28 (48 months in total). For each monitoring site, there are 11 continuous time series variables measured (e.g. PM2.5, PM10, SO2). We aggregate variables from 12 sites together so this dataset has 132 features. There are a total of 1.6\% missing values in this dataset. The test set takes data from the first 10 months (2013/03 - 2013/12). The validation set contains data from the following 10 months (2014/01 - 2014/10). The training set takes the left 28 months (2014/11 - 2017/02). To generate time series data samples, we select every 24 hours data, i.e. every 24 consecutive steps, as one sample. Similar to dataset PhysioNet-2012, 10\% observed values in the validation set and test set are eliminated and held out as ground-truth for evaluation.

\textbf{Electricity Load Diagrams (Electricity)} \hspace{1em} 
This is another widely-used public dataset from UCI~\cite{Dua2017UCI}. It contains electricity consumption data (in kWh) collected from 370 clients every 15 minutes and has no missing data. The period of this dataset is from 2011/01/01 to 2014/12/31 (48 months in total). Similar to processing Air-Quality, we use the first 10 months of data (2011/01 - 2011/10) as the test set, the following 10 months of data (2011/11 - 2012/08) as the validation set and the left (2012/09 - 2014/12) as the training set. Every 100 consecutive steps are selected as a sample to generate time-series data for model training. Due to this dataset having no missing values, we vary artificial missing rate from 10\% $\sim$ 90\% to eliminate observed values in the training set, validation set, and test set. This can make the comparison between our method and other SOTA models more comprehensive. Artificial missing values in the validation and test set are held out for model evaluation. Experiment results of 10\% missing values are displayed in Table~\ref{tb1}. Results of 20\% $\sim$ 90\% missing values are shown in Table~\ref{full_tb2}.

\textbf{Electricity Transformer Temperature (ETT)} \hspace{1em}
This dataset~\cite{Zhou2021informer} is collected from electricity transformers in two years from 2016/07/01 to 2018/06/26. Here we select the 15-minute-level version dataset that contains data collected every 15 minutes and has a total of 69680 sample points without missing values. Each sample has seven features, including oil temperature and six different types of external power load features. The data in the first four months (2016/07 - 2016/10) is taken as the test set. The following four-month data (2016/11 - 2017/02) is held out as the validation set. The left sixteen months (2017/03 - 2018/06) are for training use. The sliding-window method is applied for time-series sample generation. The window size is set as 6 hours (i.e. 24 consecutive steps) and the sliding size is 3 hours (i.e. 12 steps). Still, 10\% of the observed values in the validation set and test set are randomly masked for model evaluation.

\subsection{Baseline Methods}
To obtain a thorough comparison, we compare our method with two naive imputation methods and five recent SOTA deep learning models: 
(1). \textbf{Median}: missing values are filled with corresponding median values from the training set; (2). \textbf{Last}: in each sample, missing values are filled by the last previous observations of given features, and 0 will be filled in if there is no previous observation, ; (3). \textbf{GRUI-GAN}~\footnote{\url{https://github.com/Luoyonghong/Multivariate-Time-Series-Imputation-with-Generative-Adversarial-Networks}}~\cite{Luo2018GRUI}; (4). \textbf{E$^2$GAN}~\footnote{\url{https://github.com/Luoyonghong/E2EGAN}}~\cite{Luo2019E2GAN}; (5). \textbf{M-RNN}~\footnote{\url{https://github.com/jsyoon0823/MRNN}}~\cite{Yoon2019MRNN}; (6). \textbf{GP-VAE}~\footnote{\url{https://github.com/ratschlab/GP-VAE}}~\cite{Fortuin2020GPVAE}; (7). \textbf{BRITS}~\footnote{\url{https://github.com/caow13/BRITS}}~\cite{Cao2018BRITS}. The deep learning models have already been introduced in Section~\ref{related_work}.

\subsection{Experimental Setup}	\label{experimental_setup}
Three metrics are utilized to evaluate the imputation performance of methods: MAE (Mean Absolute Error), RMSE (Root Mean Square Error), and MRE (Mean Relative Error). The math definitions of three evaluation metrics are presented below. Note that errors are only computed on the values indicated by $mask$ in the input of the equations.

\begin{align*}
	\text{MAE}\left(estimation, target, mask \right) &= \frac{\sum_{d=1}^D \sum_{t=1}^T \lvert \left(estimation - target \right)\odot mask\rvert_t^d}{\sum_{d=1}^D \sum_{t=1}^T mask_t^d} \\
	\text{RMSE}\left(estimation, target, mask \right)&= \sqrt{\frac{\sum_{d=1}^D \sum_{t=1}^T \left(\left(\left(estimation - target\right) \odot mask \right)^2\right)_t^d}{\sum_{d=1}^D \sum_{t=1}^T mask_t^d}} \\
	\text{MRE}\left(estimation, target, mask \right) &= \frac{\sum_{d=1}^D \sum_{t=1}^T \lvert \left( estimation - target \right) \odot mask \rvert_t^d}{\sum_{d=1}^D \sum_{t=1}^T |target \odot mask |_t^d}
\end{align*}

The batch size is fixed as 128, and the early stopping strategy is applied in the model training. Training of models is stopped after 30 epochs without any decrease of MAE. To permit a fair comparison between the models, the hyper-parameter searches are executed for every model on each dataset, except SAITS-base. For SAITS-base, we fix its hyper-parameters to form a base model and observe its performance. SAITS-base is also applied in ablation experiments in Section~\ref{ablation experiments} as a baseline to make the comparisons more straightforward. Please consult Appendix~\ref{appendix: hyper-parameter search} for further details of models' hyper-parameters. Transformer used in this paper only includes the encoder part because the imputation problem is not treated as a generative task in this work, therefore, the decoder part is in no need. All models are trained with the Adam optimizer~\cite{Kingma2015Adam} on \textit{Nvidia Quadro RTX 5000} GPUs. Our models are implemented with PyTorch~\cite{Paszke2019PyTorch}.

\subsection{Experimental Results} \label{experimental_results}
The adequate experiments performed to benchmark the performance of SAITS are made up of three parts in this section. In Subsection~\ref{imputation_acc_comp}, the baseline methods and the self-attention models are impartially compared on four datasets. To further discuss the influence of imputation quality on pattern recognition tasks, an experiment in Subsection~\ref{downstream_task} is performed on a downstream classification task. In addition, we experiment to compare SAITS with NRTSI, another SOTA self-attention-based imputation model for time series. Due to bugs in the official implementation of NRTSI, it is not appropriate to put it together with other baseline methods. Hence, we run SAITS on preprocessed datasets provided by authors of NRTSI to make a fair comparison in Subsection~\ref{Comp_with_NRTSI}.

\subsubsection{Imputation Performance Comparison} \label{imputation_acc_comp}
\begin{table}[!htb]
	\caption{Performance comparison between methods on four datasets. 10\% observations in the test set are held out for evaluation. Metrics are reported in the order of MAE / RMSE / MRE. The lower, the better. Bold font indicates the best performance. GRUI-GAN and E$^2$GAN have no results on Electricity because they fail in the training due to loss explosion.}
	\label{tb1}
	\centering
	\begin{minipage}{1\textwidth}
	\resizebox{163mm}{!}{
	\begin{tabular}{p{50pt}<{\centering}|p{90pt}<{\centering}|p{90pt}<{\centering}|p{90pt}<{\centering}|p{90pt}<{\centering}}
		\toprule
		Method       & PhysioNet-2012            & Air-Quality                              & Electricity                        &  ETT \\
		\midrule
		Median       & 0.726 / 0.988 / 103.5\%   & 0.763 / 1.175 / 107.4\% & 2.056 / 2.732 / 110.1\%  & 1.145 / 1.847 / 139.1\%       \\
		\midrule
		Last         & 0.862 / 1.207 / 123.0\%   & 0.967 / 1.408 / 136.3\% & 1.006 / 1.533 / 53.9\%       &  1.007 / 1.365 / 96.4\%    \\
		\midrule
		GRUI-GAN     & 0.765 / 1.040 / 109.1\%   & 0.788 / 1.179 / 111.0\% & /                                           & 0.612 / 0.729 / 95.1\% \\
		\midrule
		E$^2$GAN     & 0.702 / 0.964 / 100.1\%   & 0.750 / 1.126 / 105.6\% & /                                         & 0.584 / 0.703 / 89.0\% \\
		\midrule
		M-RNN        & 0.533 / 0.776 / 76.0\%    & 0.294 / 0.643 / 41.4\%  & 1.244 / 1.867 / 66.6\%     & 0.376 / 0.428 / 31.6\%   \\
		\midrule
		GP-VAE       & 0.398 / 0.630 / 56.7\%    & 0.268 / 0.614 / 37.7\%  & 1.094 / 1.565 / 58.6\%    & 0.274 / 0.307 / 15.5\%  \\
		\midrule
		BRITS        & 0.256 / 0.767 / 36.5\%    & 0.153 / 0.525 / 21.6\%  & 0.847 / 1.322 / 45.3\%        & 0.130 / 0.259 / 12.5\% \\
		\midrule
		Transformer  & 0.190 / 0.445 / 26.9\%    & 0.158 / 0.521 / 22.3\%  & 0.823 / 1.301 / 44.0\%  & 0.114 / 0.173 / 10.9\%       \\
		\midrule
		SAITS-base & 0.192 / 0.439 / 27.3\%    & 0.146 / 0.521 / 20.6\%  & 0.822 / 1.221 / 44.0\%  &    0.121 / 0.197 / 11.6\%     \vspace{0.5em}  \\ 
		SAITS        & \textbf{0.186 / 0.431 / 26.6\%} & \textbf{0.137 / 0.518 / 19.3\%}  & \textbf{0.735 / 1.162 / 39.4\%}      &   \textbf{0.092  / 0.139  / 8.8\%}\\
		\bottomrule
	\end{tabular}
}
\end{minipage}
\end{table}

\begin{table}[!htb]
	\centering
	\caption{Models' parameter number (in million) and training time of each epoch (in seconds) on datasets PhysioNet-2012, Air-Quality, Electricity, and ETT are listed from left to right. GRUI-GAN and E$^2$GAN have no results for the Electricity dataset because they fail on this dataset due to loss explosion.}
	\label{tb: model_size_and_training_time}
	\begin{minipage}{1\textwidth}
	\resizebox{163mm}{!}{
		\begin{tabular}{p{50pt}<{\centering}|p{43pt}<{\centering}|p{35pt}<{\centering}|p{43pt}<{\centering}|p{35pt}<{\centering}|p{43pt}<{\centering}|p{35pt}<{\centering}|p{43pt}<{\centering}|p{35pt}<{\centering}}
			\toprule
			&   \multicolumn{2}{c|}{PhysioNet-2012}	& \multicolumn{2}{c|}{Air-Quality} &	\multicolumn{2}{c|}{Electricity}  &	\multicolumn{2}{c}{ETT}\\
			\midrule
			Model         & \# of param   & s / epoch  & \# of param       & s / epoch  & \# of param   & s / epoch  & \# of param   & s / epoch  \\
			\midrule
			GRUI-GAN      & 0.16M      & 14.4        & \phantom{0}2.32M  & 2.0      &	/           &  /        &   1.52M  &  \phantom{0}9.7\\
			\midrule
			E$^2$GAN      & 0.08M      & 22.8        & \phantom{0}1.13M   & 2.2     &	/      &  /           &    0.29M   & 11.9    \\
			\midrule
			M-RNN         & 0.07M      & \phantom{0}6.8 & \phantom{0}1.09M   & 1.3  & 18.63M     & \phantom{00}3.9      &0.15M & \phantom{0}7.6  \\
			\midrule
			GP-VAE        & 0.15M      & 40.1         & \phantom{0}0.36M   & 8.7     & 13.45M	   & 106.0        &           0.27M      & 22.1\\
			\midrule
			BRITS         & 0.73M      & 12.8         & 11.25M              & 1.9    &\phantom{0}7.00M   & \phantom{00}5.2                       & 0.57M   & 10.2 \\
			\midrule
			Transformer   & 4.36M      & \phantom{0}3.1  & \phantom{0}5.13M   & 0.9  & 14.78M     & \phantom{00}2.6           &  4.67M & \phantom{0}3.7  \\
			\midrule
			SAITS-base  & 1.38M   & \phantom{0}2.7 & \phantom{0}1.56M  & 1.1 & \phantom{0}2.20M &  \phantom{00}2.1   &  1.33M  & \phantom{0}2.5  \vspace{0.5em} \\  
			SAITS         & 5.32M      & \phantom{0}5.0  & \phantom{0}3.07M   & 0.9   & 11.51M & \phantom{00}2.6                    &   4.27M & \phantom{0}4.6  \\
			\bottomrule
		\end{tabular}
	}
\end{minipage}
\end{table}
\begin{table}[!htb]
	\centering
	\caption{Performance comparison between methods on the Electricity dataset across different missing rates from 20\% $\sim$ 90\%. Metrics are reported in the order of MAE / RMSE / MRE. The lower, the better. Values in bold are the best.}
	\label{full_tb2}
	\begin{minipage}{1\textwidth}
		\resizebox{163mm}{!}{
			\begin{tabular}{p{60pt}<{\centering}|p{90pt}<{\centering}|p{90pt}<{\centering}|p{90pt}<{\centering}|p{90pt}<{\centering}}
				\toprule
				Method         & 20\%                     & 30\%                    & 40\%                   & 50\%                    \\
				\midrule
				Median        & 2.053 / 2.726 / 109.9\%  & 2.055 / 2.732 / 110.0\% & 2.058 / 2.734 / 110.2\%   & 2.053 / 2.728 / 109.9\% \\
				\midrule
				Last          & 1.012 / 1.547 / 54.2\%   & 1.018 / 1.559 / 54.5\%  & 1.025 / 1.578 / 54.9\%  & 1.032 / 1.595 / 55.2\%   \\
				\midrule
				M-RNN         & 1.242 / 1.854 / 66.5\%   & 1.258 / 1.876 / 67.3\%  & 1.269 / 1.884 / 68.0\%  & 1.283 / 1.902 / 68.7\%     \\
				\midrule
				GP-VAE        & 1.124 / 1.502 / 60.2\%   & 1.057 / 1.571 / 56.6\%  & 1.090 / 1.578 / 58.4\% & 1.097 / 1.572 / 58.8\%   \\
				\midrule
				BRITS         & 0.928 / 1.395 / 49.7\%   & 0.943 / 1.435 / 50.4\%  & 0.996 / 1.504 / 53.4\%  & 1.037 / 1.538 / 55.5\%   \\
				\midrule
				Transformer   & 0.843 / 1.318 / 45.1\%   & 0.846 / 1.321 / 45.3\%  & 0.876 / 1.387 / 46.9\% & 0.895 / 1.410 / 47.9\%   \\
				\midrule
				SAITS-base & 0.838 / 1.264 / 44.9\%   & 0.845 / 1.247 / 45.2\%  & 0.873 / 1.325 / \textbf{46.7\%} & 0.939 / 1.537 / 50.3\%  \vspace{0.5em} \\ 
				SAITS  &\textbf{0.763 / 1.187 / 40.8\%} &\textbf{0.790 / 1.223 / 42.3\%}   & \textbf{0.869} / \textbf{1.314 / 46.7\%}  & \textbf{0.876 / 1.377 / 46.9\%}  \\
				\bottomrule
			\end{tabular}
		}
	\end{minipage}
	\begin{minipage}{1\textwidth}
		\resizebox{163mm}{!}{
			\begin{tabular}{p{60pt}<{\centering}|p{90pt}<{\centering}|p{90pt}<{\centering}|p{90pt}<{\centering}|p{90pt}<{\centering}}
				\toprule
				Method         & 60\%                    & 70\%                    & 80\%                     & 90\%                    \\
				\midrule
				Median         & 2.057 / 2.734 / 110.2\% & 2.050 / 2.726 / 109.8\% & 2.059 / 2.734 / 110.2\%   & 2.056 / 2.723 / 110.1\% \\
				\midrule
				Last          & 1.040 / 1.615 / 55.7\%  & 1.049 / 1.640 / 56.2\%  & 1.059 / 1.663 / 56.7\%    & 1.070 / 1.690 / 57.3\% \\
				\midrule
				M-RNN         & 1.298 / 1.912 / 69.4\%  & 1.305 / 1.928 / 69.9\%   & 1.318 / 1.951 / 70.5\%    & 1.331 / 1.961 / 71.3\%  \\
				\midrule
				GP-VAE         & 1.101 / 1.616 / 59.0\%  & 1.037 / 1.598 / 55.6\%  & 1.062 / 1.621 / 56.8\%    & 1.004 / 1.622 / 53.7\%     \\
				\midrule
				BRITS          & 1.101 / 1.602 / 59.0\%   & 1.090 / 1.617 / 58.4\%  & 1.138 / 1.665 / 61.0\%    & 1.163 / 1.702 / 62.3\%  \\
				\midrule
				Transformer    & \textbf{0.891} / 1.404 / \textbf{47.7\%}  & 0.920 / 1.437 / 49.3\%  & 0.924 / 1.472 / 49.5\%    & 0.934 / 1.491 / \textbf{49.8\%}  \\
				\midrule
				SAITS-base  & 0.969 / 1.565 / 51.9\%  & 0.972 / 1.601 / 52.0\% & 1.012 / 1.608 / 54.2\%     & 1.001 / 1.630 / 53.6\%\vspace{0.5em} \\ 
				SAITS          & 0.892 / \textbf{1.328} / 47.9\%  & \textbf{0.898 / 1.273 / 48.1\%}  &\textbf{0.908 / 1.327 / 48.6\%}& \textbf{0.933} / \textbf{1.354} / 49.9\%  \\
				\bottomrule
			\end{tabular}
		}
	\end{minipage}
\end{table}

Table~\ref{tb1} reports the imputation performance of models on four datasets in three evaluation metrics (MAE / RMSE / MRE). On PhysioNet-2012 and Air-Quality, GRUI-GAN achieves better results than Last but is slightly inferior to Median. E$^2$GAN performs better than these three methods. On Electricity, GRUI-GAN and E$^2$GAN both fail because of loss explosion. We find this is caused by the long sequence length. Dataset Electricity's sequence length is 100. If the sequence length of the Air-Quality dataset is increased from 24 to 100, both GAN models will be confronted with loss explosion and fail again. M-RNN outperforms both naive imputation methods a lot on PhysioNet-2012 and Air-Quality but gets worse results than Last on Electricity. On ETT, all the deep-learning methods performs obviously better than the naive imputation methods. GP-VAE and BRITS both perform much better than the methods mentioned above on all the four datasets. BRITS is the best one among baseline methods. When it comes to the self-attention-based models, Transformer surpasses BRITS obviously on datasets PhysioNet-2012, Electricity, and ETT, and obtains comparable results to BRITS on Air-Quality. SAITS-base achieves similar results to Transformer on all datasets. SAITS exceeds all baseline methods significantly on all metrics and all datasets, and it outperforms Transformer and SAITS-base as well.

To show further details of the models in Table~\ref{tb1}, the parameter number and the training speed of them are listed in the following Table~\ref{tb: model_size_and_training_time}. We can see that GP-VAE is the slowest model and consumes the most seconds for each epoch training. The RNN-based models are all slower than the self-attention-based models. Compared to BRITS that yields the best results in the baseline methods, SAITS takes half the training time or even less as BRITS on each epoch. Compared to Transformer, SAITS-base has only 15\% $\sim$ 30\% parameters of Transformer's, but it still obtains comparable performance to Transformer. It confirms that SAITS' model structure is more efficient than Transformer on the time-series imputation task.

To further compare the performance of the imputation methods on different missing rates, we also experiment to introduce missing values into the Electricity dataset at different rates between 20\% $\sim$ 90\%. The results of this experiment are elaborated in Table~\ref{full_tb2}. The results of 10\% missing rate have been displayed in Table~\ref{tb1}. Dataset Electricity is selected because it has no missing data, and it is the most complex among the four datasets because each sample has 370 features and 100 time steps (please refer to Table~\ref{dataset_info}). In Table~\ref{tb1}, the models achieve the highest error on the Electricity dataset, which also proves that it is the most difficult one to impute among four datasets. Therefore, Electricity is the most suitable dataset to experiment with different missing rates. GRUI-GAN and E$^2$GAN are omitted because they fail on the Electricity dataset due to loss explosion as is discussed above. The baseline methods are all inferior to self-attention-based models in all cases. SAITS-base performs better than Transformer in cases of 20\%, 30\%, and 40\%. However, its performance becomes worse than Transformer in the left cases where missing rates become higher. This is because the hyper-parameters of SAITS-base are fixed, and its number of parameters is limited to a low level, only 15\% of Transformer (refer to Table~\ref{tb: model_size_and_training_time}). Such a situation makes the capacity of SAITS-base not enough to well handle the imputation problem with the higher missing rate. And given enough model capacity (with 78\% parameters of Transformer), SAITS achieves the best performance in eight out of nine cases, demonstrating its distinct advantage.

\subsubsection{Downstream Classification Task} \label{downstream_task}
In the PhysioNet-2012 dataset, each sample has a label indicating if the patient is deceased that makes PhysioNet-2012 a binary-classification dataset and there are 1,707 (14.2\%) samples with the positive mortality label. Consequently, mortality prediction is one of the main tasks on this dataset. However, 80\% missing values make this task challenging. Similar to PhysioNet-2012, in real-world datasets, missingness often makes tasks of pattern recognition tricky. To further discuss the benefits that SAITS can bring to pattern recognition, the experiment here is conducted on a downstream classification task on the PhysioNet-2012 dataset to qualitatively compare the imputation quality of each method. Note that this experiment is inspired by prior work~\cite{Cao2018BRITS, Luo2018GRUI, Luo2019E2GAN, Fortuin2020GPVAE, Ramchandran2021LVAE}. The idea behind this experimental design is that, if one method's imputation is better in terms of overall data quality, datasets imputed by the method should achieve better performance on downstream pattern recognition tasks, such as classification here.

\begin{table}[!htb]
	\caption{Results of the downstream classification task on the PhysioNet-2012 dataset. Performance metrics of methods are calculated by five independent runs. The reported values are means $\pm$ standard deviations. The higher, the better. Values in bold are the best.}
	\label{tb3}
	\centering
	\begin{tabular}{p{70pt}<{\centering}|p{90pt}<{\centering}|p{90pt}<{\centering}|p{90pt}<{\centering}}
		\toprule
		Method     & ROC-AUC           & PR-AUC              & F1-score              \\
		\midrule
		Median     & 83.4\% $\pm$ 0.4\% & 46.0\% $\pm$ 0.6\%  & 38.5\% $\pm$ 3.1\%  \\
		\midrule
		Last       & 82.8\% $\pm$ 0.3\% & 46.9\% $\pm$ 0.4\%  & 39.5\% $\pm$ 2.4\%  \\
		\midrule
		GRUI-GAN   & 83.0\% $\pm$ 0.2\% & 45.1\% $\pm$ 0.7\%  & 38.8\% $\pm$ 2.0\% \\
		\midrule
		E$^2$GAN   & 83.0\% $\pm$ 0.2\% & 45.5\% $\pm$ 0.5\%  & 35.6\% $\pm$ 2.0\%  \\
		\midrule
		M-RNN      & 82.2\% $\pm$ 0.2\% & 45.4\% $\pm$ 0.6\%  & 38.8\% $\pm$ 3.5\% \\
		\midrule
		GP-VAE     & 83.4\% $\pm$ 0.2\% & 48.1\% $\pm$ 0.7\%  & 40.9\% $\pm$ 3.3\% \\
		\midrule
		BRITS      & 83.5\% $\pm$ 0.1\% & 49.1\% $\pm$ 0.4\%  & 41.3\% $\pm$ 1.8\% \\
		\midrule
		Transformer& 84.3\% $\pm$ 0.5\% & 49.2\% $\pm$ 1.4\%  & 41.2\% $\pm$ 1.9\% \\
		\midrule
		SAITS-base& 84.6\% $\pm$ 0.2\% & 49.8\% $\pm$ 0.4\% & 41.5\% $\pm$ 2.0\%\vspace{0.5em} \\ 
		SAITS      & \textbf{84.8\% $\pm$ 0.2\%} & \textbf{51.0\% $\pm$ 0.5\%}  & \textbf{42.7\% $\pm$ 2.8\%} \\
		\bottomrule
	\end{tabular}
\end{table}

We first let each method impute the dataset and then train a classifier on each imputed dataset to obtain the classification results. Since this is a time-series dataset, a simple RNN classification model is employed as the classifier. This RNN classifier consists of a GRU layer followed by a fully connected layer. All hyper-parameters are fixed as follows: the learning rate ($1 \times 10^{-3}$), the batch size (128), the RNN hidden size (128), the patience of the early stopping strategy (20). The classifier and the training procedure are kept exactly the same for each imputed dataset to obtain the equitably comparable results. Considering that classes in this dataset are imbalanced, metrics ROC-AUC (Area Under ROC Curve), PR-AUC (Area Under Precision-Recall Curve), and F1-score are used to measure performance. The experiment results are reported in Table~\ref{tb3}. The method names in the table annotate that the dataset is imputed by which method. 

As displayed in Table~\ref{tb3}, the classifier trained on the dataset imputed by SAITS achieves the greatest results on all evaluation metrics and obtains obvious improvements than RNN-based models (1.3\%, 1.9\%, and 1.4\% better than BRITS in ROC-AUC, PR-AUC, and F1-score). Even though comparing SAITS with Transformer, the improvements on PR-AUC and F1-score are noteworthy (increased 1.8\% and 1.5\% respectively), considering this is on an imbalanced and sparse dataset and the classifier is trivial. Such an indirect comparison reaches the same conclusion that the imputation quality of SAITS is the best among all methods. In the meanwhile, it demonstrates that SAITS does not only have higher accuracy in the imputation metrics but also can help improve the performance of trivial models on pattern recognition tasks on time-series datasets with missing values.

\subsubsection{Comparison with NRTSI} \label{Comp_with_NRTSI}
To make a fair comparison with NRTSI~\cite{Shan2021NRTSI}, another SOTA self-attention-based imputation model for time series, we tried to incorporate it into our framework of imputation models. However, its official open-source implementation in the GitHub repository \url{https://github.com/lupalab/NRTSI} has fatal bugs that stop us from reproducing their results or adding their model into our baselines. Fortunately, the authors provide their preprocessed datasets Air and Gas in their code repository. This makes us able to run SAITS on their datasets for an impartial comparison with NRTSI. Regarding the experiments in this section, we have uploaded the related dataset generating script and SAITS configuration files to our code repository to ensure our results are reproducible.

Both datasets Air~\cite{Zhang2017AirQuality} and Gas~\cite{Burgues2018Gas} are from UCI machine learning repository~\cite{Dua2017UCI}. The raw data of the Air dataset is the same as the Air-Quality dataset used in our work, but the preprocessing methods are different that results in different time steps and feature numbers. Our Air-Quality dataset is 132-dimensional and has 24 time steps, while the Air dataset is 11-dimensional and has 48 time steps. The experiment results on the datasets Air and Gas are listed in Table~\ref{comp_NRTSI_air} and~\ref{comp_NRTSI_gas} respectively. The results of NRTSI are from Table 5 in the original paper~\cite{Shan2021NRTSI}, where the evaluation metric is mean squared error (MSE), so we keep using it here.

\begin{table} [!htb]
	\centering
	\caption{The performance comparison between NRTSI and SAITS on dataset Air across different missing rates from 10\% $\sim$ 80\%. The evaluation metric is MSE. The lower, the better. The best results are in bold.}
	\label{comp_NRTSI_air}
	\begin{tabular}{p{55pt}<{\centering}|p{35pt}<{\centering}|p{35pt}<{\centering}|p{35pt}<{\centering}|p{35pt}<{\centering}|p{35pt}<{\centering}|p{35pt}<{\centering}|p{35pt}<{\centering}|p{35pt}<{\centering}}
		\toprule
		Method        & 10\%     & 20\%      & 30\%     & 40\%      & 50\%     & 60\%      & 70\%     & 80\%     \\
		\midrule
		NRTSI        & 0.1230   & 0.1155    & 0.1189   & 0.1250    & 0.1297   & 0.1378    & 0.1542   & 0.1790   \\
		\midrule
		SAITS         & \textbf{0.0980}   & \textbf{0.0911}    & \textbf{0.0916}   & \textbf{0.1021}    & \textbf{0.1109}   & \textbf{0.1030}    & \textbf{0.1179}   & \textbf{0.1409} \\
		\bottomrule
	\end{tabular}
\end{table}

\begin{table} [!htb]
	\centering
	\caption{The performance comparison between NRTSI and SAITS on dataset Gas across different missing rates from 10\% $\sim$ 80\%. The evaluation metric used here is MSE. The lower, the better. The best results are in bold.}
	\label{comp_NRTSI_gas}
	\begin{tabular}{p{55pt}<{\centering}|p{35pt}<{\centering}|p{35pt}<{\centering}|p{35pt}<{\centering}|p{35pt}<{\centering}|p{35pt}<{\centering}|p{35pt}<{\centering}|p{35pt}<{\centering}|p{35pt}<{\centering}}
		\toprule
		Method        & 10\%     & 20\%      & 30\%     & 40\%      & 50\%     & 60\%      & 70\%     & 80\%     \\
		\midrule
		NRTSI         & 0.0165   & 0.0195    & 0.0196   & 0.0229    & 0.0286   & 0.0311    & 0.0362   & 0.0445  \\
		\midrule
		SAITS         & \textbf{0.0100}   & \textbf{0.0123}    & \textbf{0.0145}   & \textbf{0.0192}    & \textbf{0.0239}   & \textbf{0.0289}    & \textbf{0.0337}   & \textbf{0.0326} \\
		\bottomrule
	\end{tabular}
\end{table}
\begin{figure}[!htb]
	\centering
	\subfigure[Comparison on dataset Air]{
		\begin{minipage}[!t]{0.48\textwidth}
			\centering
			\includegraphics[width=8cm]{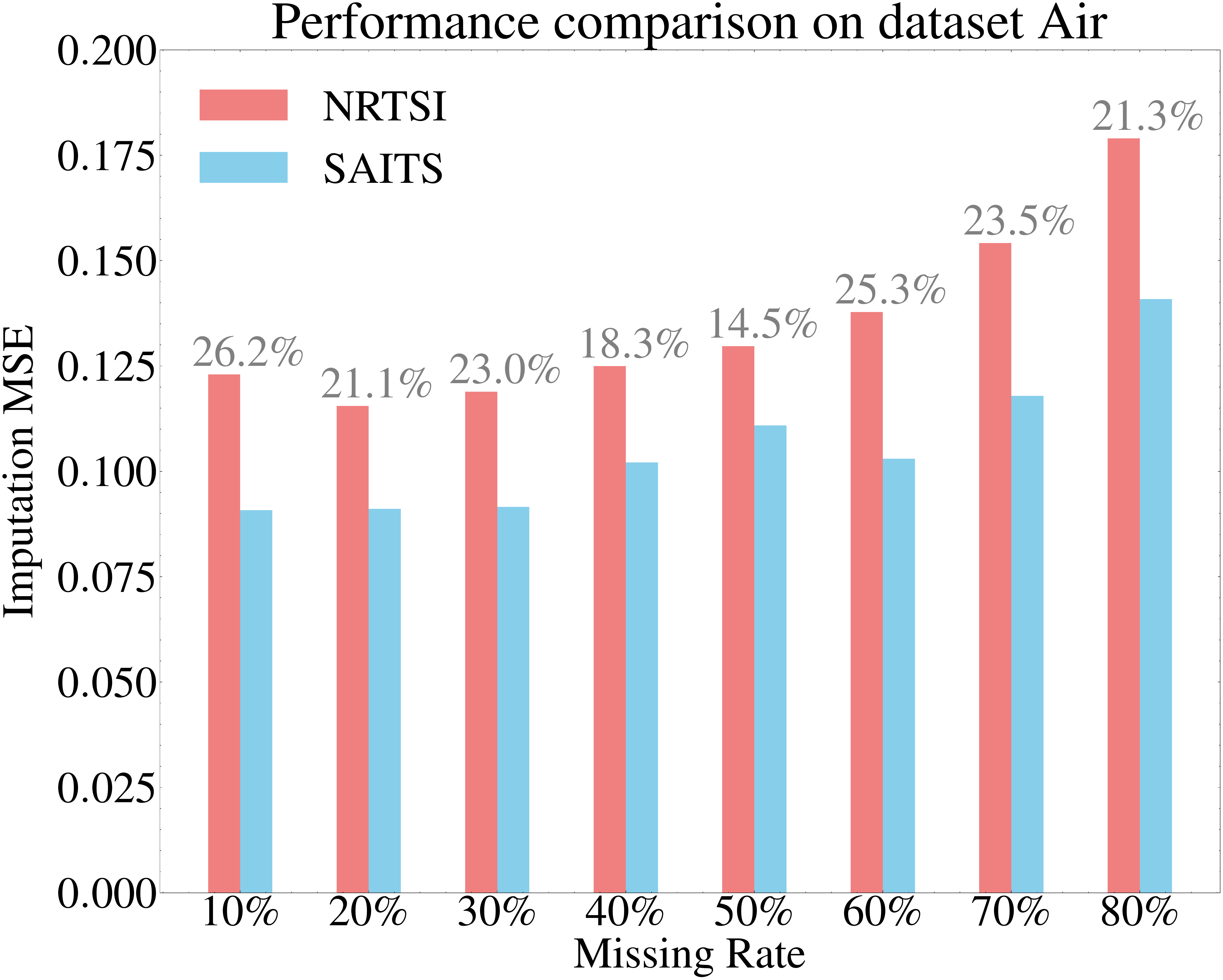}
		\end{minipage}
	}
	\subfigure[Comparison on dataset Gas]{
		\begin{minipage}[!t]{0.48\textwidth}
			\centering
			\includegraphics[width=8cm]{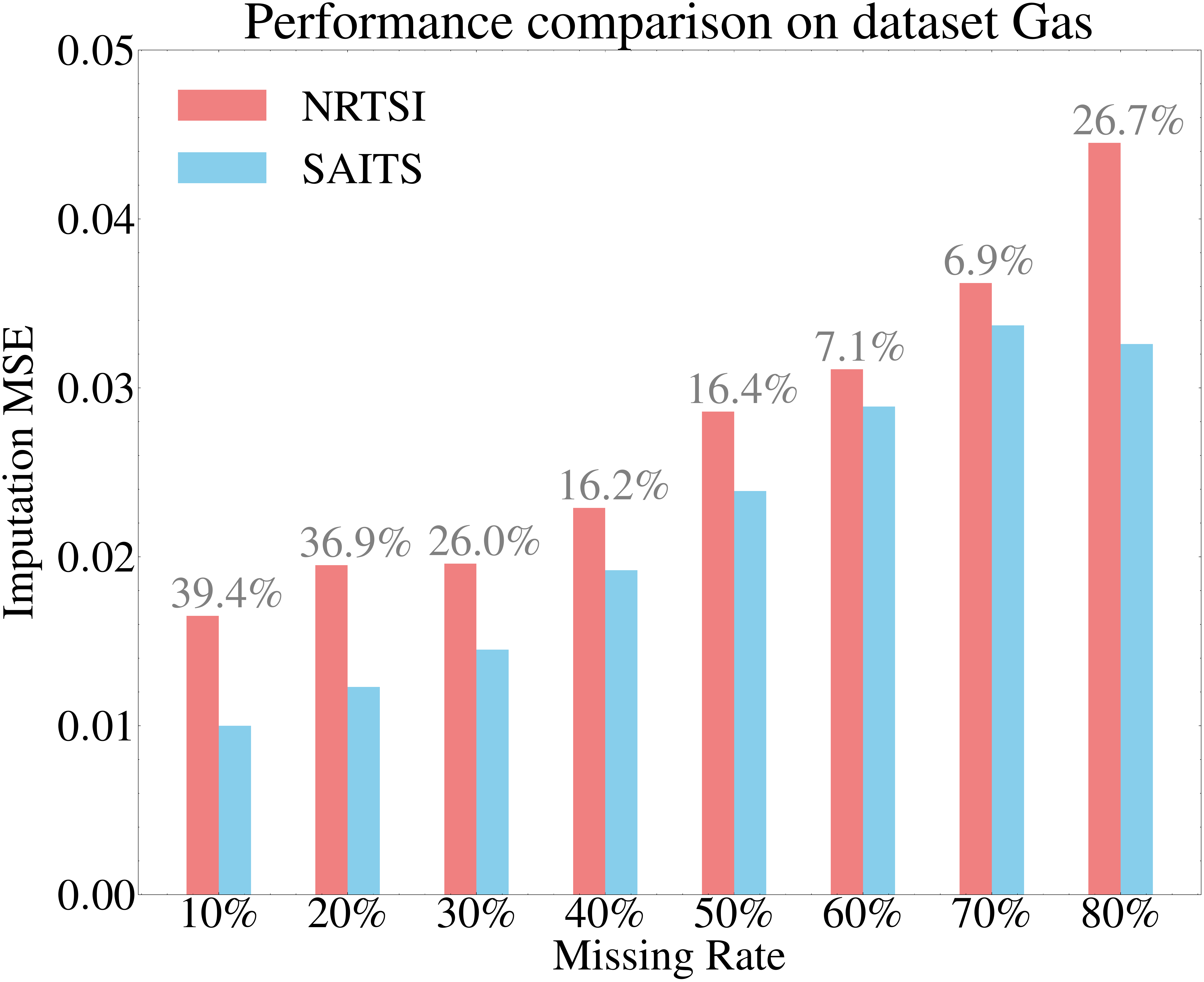}
		\end{minipage}
	}
	\caption{The visualized comparison with NRTSI on datasets Air and Gas. The percentage numbers above the bars indicate, compared with NRTSI, the amount of imputation MSE reduced by SAITS.}
	\label{fig:NRTSI_VS_SAITS}
\end{figure}

With the above results, it is obvious that SAITS outperforms NRTSI in all cases on both datasets. In particular, SAITS achieves 7\% $\sim$ 39\% smaller MSE (above 20\% in nine out of sixteen cases) than NRTSI. To make the comparison more straightforward, the bar graphs in Figure~\ref{fig:NRTSI_VS_SAITS} are plotted to visualize the results in Table~\ref{comp_NRTSI_air} and~\ref{comp_NRTSI_gas}.

\begin{table}[!htb]
	\centering
	\caption{The models' parameter number (in million) and training time of each epoch (in seconds) on datasets Air and Gas are listed below. NRTSI have no results of the training speed because the original paper does not include them.}
	\label{tb:NRTSI_SAITS_param_speed}
	\begin{tabular}{p{50pt}<{\centering}|p{60pt}<{\centering}|p{60pt}<{\centering}|p{60pt}<{\centering}|p{60pt}<{\centering}}
		\toprule
		&   \multicolumn{2}{c|}{Air}	& \multicolumn{2}{c}{Gas}		\\
		\midrule
		Model         & \# of param   & s / epoch  & \# of param   & s / epoch  \\
		\midrule
		NRTSI         & 84.00M        &  /       &  84.00M       &   /       \\
		\midrule
		SAITS         & 10.00M        & 2.6      &  \phantom{0}2.78M   & 18.4 \\
		\bottomrule
	\end{tabular}
\end{table}

Besides model performance, the model parameter numbers and training speed are recorded in Table~\ref{tb:NRTSI_SAITS_param_speed}. NRTSI's number of parameters is from Appendix B in~\cite{Shan2021NRTSI}. The results in Table~\ref{tb:NRTSI_SAITS_param_speed} tell us that SAITS needs much fewer parameters than NRTSI on both datasets (only 12\% and 3\% of NRTSI's parameters on datasets Air and Gas respectively). Considering the hyper-parameters of NRTSI share across the datasets and do not get adjusted accordingly in the original paper~\cite{Shan2021NRTSI}, NRTSI may not need so many parameters to obtain such performance. Nevertheless, NRTSI directly takes a Transformer encoder as the backbone, and it has been proven in the experiments and discussion in Section~\ref{imputation_acc_comp} that SAITS architecture is more efficient than Transformer on the imputation task. Concerning the training speed, we can not run NRTSI, so NRTSI's speed does not have records. However, according to the algorithms of NRTSI listed in Appendix A in~\cite{Shan2021NRTSI}, both the training procedure and imputation procedure have two nested loops, which can make NRTSI much slower than SAITS because SAITS has no loop in neither training nor imputation stage.

\subsection{Ablation Studies} \label{ablation experiments}
In this section, three ablation experiments are leveraged to discuss the rationality of SAITS architecture design. The first one~\ref{diagonal_mask_ablation_study} is to validate the improvement brought by the diagonally-masked self-attention (DMSA). The second one~\ref{weighted_combination_ablation_study} is to discuss the necessity of the weighted combination. The third one~\ref{why_not_more_than_2_blocks} is to explain why we do not apply more than two DMSA blocks. 

\subsubsection{Ablation Study of the Diagonal Masks in Self-Attention} \label{diagonal_mask_ablation_study}
\begin{table} [!htb]
	\caption{Ablation experiment results of the diagonal masks in self-attention. SAITS-base-w/o is the exact same as SAITS-base, except it is without the diagonal masks in self-attention layers.}
	\label{ablation_diagonal_mask}
	\centering
	\begin{minipage}{1\textwidth}
	\resizebox{163mm}{!}{
	\begin{tabular}{p{68pt}<{\centering}|p{90pt}<{\centering}|p{90pt}<{\centering}|p{90pt}<{\centering}|p{90pt}<{\centering}}
		\toprule
		Model            & PhysioNet-2012            & Air-Quality                                 & Electricity                      &   ETT    \\
		\midrule
		SAITS-base-w/o & 0.200 / 0.446 / 28.5\%    & 0.148 / 0.528 / 21.3\%  & 0.898 / 1.504 / 48.1\%   &  0.147 / 0.211 / 13.8\% \\ 
		\midrule
		SAITS-base     &\textbf{0.192 / 0.439 / 27.3\%}&\textbf{0.146 / 0.521 / 20.6\%}&\textbf{0.822 / 1.221 / 44.0\%}  &  \textbf{0.121 / 0.197 / 11.6\%}\\ 
		\bottomrule
	\end{tabular}
}
\end{minipage}
\end{table}
To prove that DMSA has better imputation performance than the conventional self-attention, a comparison is made between SAITS-base and SAITS-base-w/o in Table~\ref{ablation_diagonal_mask}. SAITS-base-w/o is without the diagonal masks. SAITS-base outperforms SAITS-base-w/o on all datasets, and this demonstrates DMSA does improve SAITS' imputation ability.

\subsubsection{Ablation Study of the Weighted Combination} \label{weighted_combination_ablation_study}
\begin{table} [!htb]
	\caption{Ablation experiment results of the weighted combination. SAITS-base-1block does not have the second DMSA block nor the weighted-combination block, and its final representation is directly from the only DMSA block. SAITS-base-R2 directly takes \textbf{Learned Representation 2} as the final representation. In other words, it has no combination of representations. SAITS-base-residual applies a residual connection to combine \textbf{Learned Representation 1 and 2}.} 
	\label{ablation_weighted_combination}
	\centering
	\resizebox{163mm}{!}{
		\begin{tabular}{p{84pt}<{\centering}|p{90pt}<{\centering}|p{90pt}<{\centering}|p{90pt}<{\centering}|p{90pt}<{\centering}}
			\toprule
			Model            & PhysioNet-2012                               & Air-Quality             & Electricity                        & ETT      \\
			\midrule
			SAITS-base-1block	& 0.204 / 0.496 / 29.2\% & 0.178 / 0.544 / 25.1\% & 0.876 / 1.381 / 46.9\%                   &  0.149 / 0.221 / 14.1\% \\ 
			\midrule
			SAITS-base-R2  & 0.199 / 0.451 / 28.4\%  & 0.149 / 0.522 / 21.0\%  & 0.906 / 1.456 / 48.5\%                         &  0.141 / 0.203 / 13.5\%\\ 
			\midrule
			SAITS-base-residual & 0.200 / 0.477 / 28.5\%  & 0.160 / 0.527 / 22.6\%  & \textbf{0.819} / 1.223 / \textbf{43.7\%} & 0.143 / 0.207 / 13.4\%   \\ 
			\midrule
			SAITS-base     &\textbf{0.192 / 0.439 / 27.3\%}&\textbf{0.146 / 0.521 / 20.6\%}& 0.822 / \textbf{1.221} / 44.0\%  & \textbf{0.121 / 0.197 / 11.6\%}  \\ 
			\bottomrule
		\end{tabular}
	}
\end{table}
During the model design process, after applying the diagonal masks to self-attention, we further think about how to enhance the imputation ability. As a result, the second DMSA block is added to increase our model's depth and extend the learning process. Rather than simply raising the layer number of the first DMSA block that can also increase the network depth, the second DMSA block is employed as a learner to play a role of verification. Different from the first DMSA block that can only make imputation from scratch, the second DMSA block has its input containing the imputed data from the first DMSA block. Accordingly, its learning target is to verify these imputation values. However, there is no guarantee that the second DMSA block can perform better than the first one. In other words, the imputations from the second DMSA block are not necessarily better than those from the first block. For example, SAITS-base-R2 achieves better performance than SAITS-base-1block on datasets PhysioNet-2012, Air-Quality, and ETT, but performs worse on the Electricity dataset. Hence, taking imputation values from either block is not wise. Therefore, we let representations from both blocks form the final imputation together, namely in the way of the weighted combination discussed in Section~\ref{SAITS: the weighted combination block}.

We compare the weighted combination with the other two designs to discuss its rationality. As shown in Table~\ref{ablation_weighted_combination}, one is no combination, directly taking \textbf{Learned Representation 2} as the final representation, referring to SAITS-base-R2. The other is the residual combination, which combines \textbf{Learned Representation 1 and 2} by a residual connection, referring to SAITS-base-residual. Compared with the residual connection, the weighted combination design parameterizes the connection process and makes it actively assign weights for the learned representations rather than being a simple addition.

As shown in Table~\ref{ablation_weighted_combination}, SAITS-base obtains the best results on datasets PhysioNet-2012, Air-Quality, and ETT. On these three datasets, SAITS-base-residual is even inferior to SAITS-base-R2. That is to say, the residual combination makes results worse. On dataset Electricity, SAITS-base and SAITS-base-residual achieve comparable results, and both are better than SAITS-base-R2. In summary, our weighted combination is the most practical design in all of the three.

\subsubsection{Ablation Study of the Third DMSA Block} \label{why_not_more_than_2_blocks}
Similar to applying the second DMSA block to obtain better performance, theoretically, we can apply more than two DMSA blocks. However, the benefit is marginal. Taking three DMSA blocks as an example, the experiments are conducted and the results are listed in Table~\ref{ablation_the_3rd_block} above.

\begin{figure}[!htb]
	\centering
	\caption{Structure illustration of SAITS with three residual-connected DMSA blocks. The second DMSA block takes in data imputed by the first DMSA block, and the third DMSA block takes in data from the second one. The final output is from a residual connection of the representations produced by three DMSA blocks.}
	\label{fig3}
	\includegraphics[width=15cm]{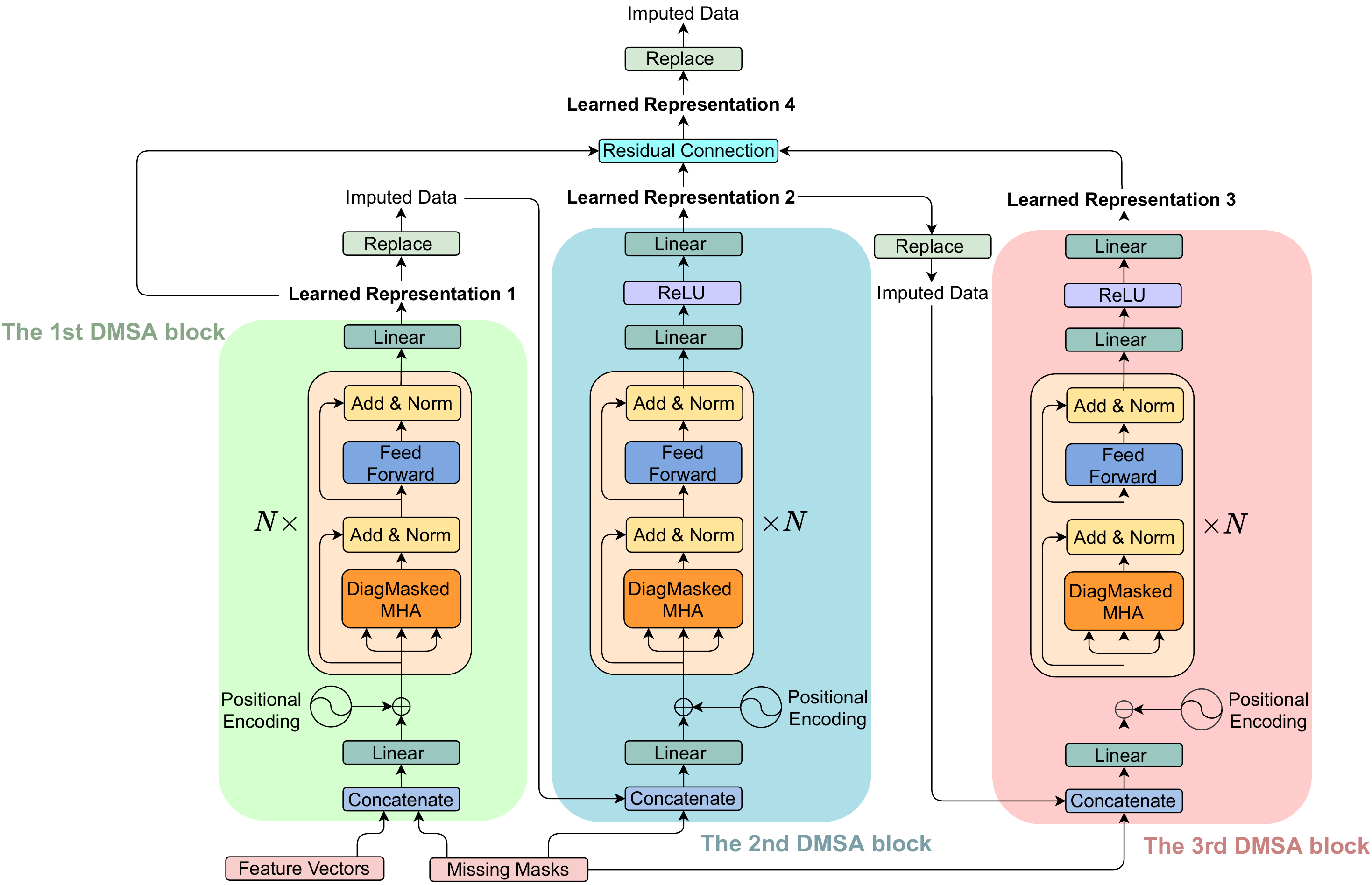}
\end{figure}

\begin{figure}[!htb]
	\caption{Structure illustration of SAITS with three cascade-weighted DMSA blocks. The representations from the first two DMSA blocks are merged by the first weighted combination block to produce Learned Representation 3, which is used to impute data for the input of the third DMSA block. The final output is a weighted combination of the representation from the third DMSA block and Learned Representation 3.}
	\centering
	\includegraphics[width=16.5cm]{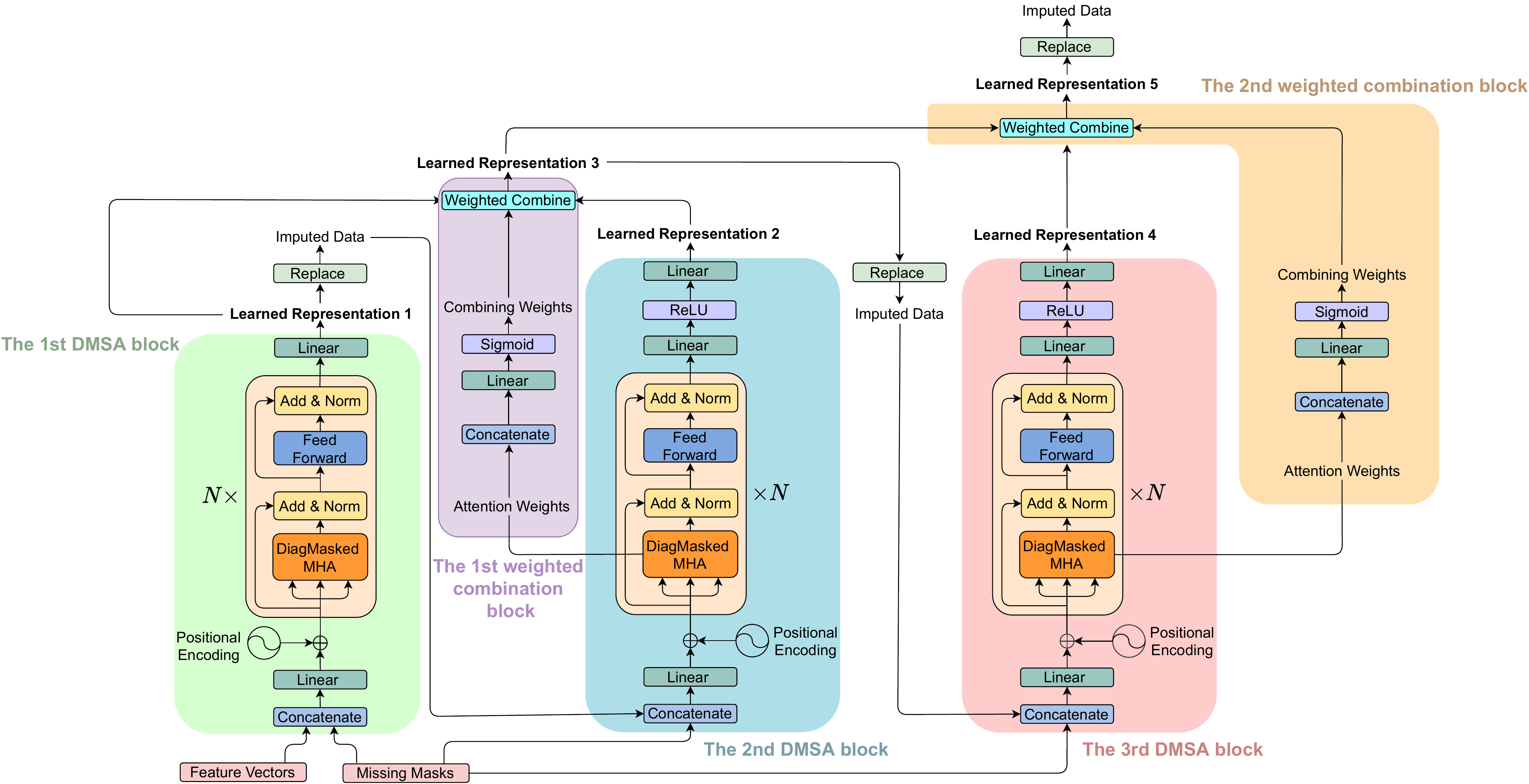}
	\label{fig4}
\end{figure}

\begin{table} [!htb]
	\caption{Ablation experiment results of the third DMSA block. Results of SAITS here are from Table~\ref{tb1} in our paper. Both SAITS-3residual and SAITS-3cascade apply the same hyper-parameters with SAITS.}
	\label{ablation_the_3rd_block}
	\centering
	\resizebox{163mm}{!}{
		\begin{tabular}{p{70pt}<{\centering}|p{90pt}<{\centering}|p{90pt}<{\centering}|p{90pt}<{\centering}|p{90pt}<{\centering}}
			\toprule
			Model            & PhysioNet-2012            & Air-Quality             & Electricity &    ETT  \\
			\midrule
			SAITS-3residual  & 0.189 / 0.620 / 27.0\% & 0.158 / \textbf{0.509} / 22.2\% & 0.740 / \textbf{1.020} / 39.6\%      &      0.103 / 0.145 / 9.6\%  \\
			\midrule
			SAITS-3cascade	& \textbf{0.185} / \textbf{0.418} / \textbf{26.4\%} & 0.146 / 0.512 / 20.5\% & 0.800 / 1.147 / 42.8\% & 0.096 / 0.141 / \textbf{8.8\%} \\
			\midrule
			SAITS           & 0.186 / 0.431 / 26.6\% & \textbf{0.137} / 0.518 / \textbf{19.3\%} & \textbf{0.735} / 1.162 / \textbf{39.4\%} & \textbf{0.092  / 0.139  / 8.8\%}\\ 
			\bottomrule
		\end{tabular}
	}
\end{table}

Regarding how to combine representations from three DMSA blocks, there are still two options: residual connection and weighted combination. Residual connection is easy to implement, and SAITS-3residual takes this way. The weighted combination can only combine two blocks' representation at a time, so the cascade-weighted combination is used here to implement SAITS-3cascade. The graphs in Figure~\ref{fig3} and Figure~\ref{fig4} are plotted to clearly illustrate both models' structure.

With the results in Table~\ref{ablation_the_3rd_block}, we can see, in general, SAITS-3residual and SAITS-3cascade do not obtain better results than SAITS, which means that adding one more block brings nothing but more parameters and computation resource waste.

\section{Conclusion} \label{conclusion}
This paper proposes SAITS, a novel self-attention-based model to impute missing values in multivariate time series. Specifically, a joint-optimization training approach is designed for self-attention-based models to perform the imputation task. Compared to BRITS, a SOTA RNN-based imputation model, SAITS reduces mean absolute error (MAE) by 12\% $\sim$ 38\% and achieves 2.0 $\sim$ 2.6 times faster training speed. In the comparison with another SOTA model NRTSI, which takes a Transformer as the backbone, SAITS achieves 7\% $\sim$ 39\% better imputation accuracy. Moreover, when Transformer is trained by our joint-optimization approach, SAITS still obtains MAE 2\% $\sim$ 19\% smaller than it, with comparable training speed. Especially on dataset Electricity, the most complex dataset among all the four, the improvement is still obvious (11\%), which means SAITS has an obvious advantage over Transformer when datasets become complex. Furthermore, the experiments also tell us that SATIS has a more efficient model structure than Transformer on the imputation task. To obtain comparable performance, SAITS needs only 15\% $\sim$ 30\% parameters of Transformer. Additionally, to justify the design of SAITS architecture, a series of ablation experiments are performed to further discuss the reasons for our design and prove its effectiveness. All of the experimental results lead to the same conclusion that SAITS efficiently achieves the new state-of-the-art accuracy on the time-series imputation task. In addition to imputation accuracy that evaluates SAITS quantitatively, our empirical results in the downstream classification experiment qualitatively show that classification performance can directly get improved by letting SAITS impute the missing part, which reveals SAITS' potential of becoming a bridge for pattern recognition models to learn with incomplete time-series data.

Our future work will investigate the imputation performance of SAITS on partially-observed time series with other missing patterns. Note that we add completely-random artificial missingness in MIT because the missing pattern is assumed to be MCAR in the settings of this work. If one already knows the missing pattern of the dataset to be imputed, one can apply the specific pattern to introduce artificially-missing values. This is intuitive and still keeps the functionality of MIT, though whether it can help improve imputation accuracy compared to applying MCAR missingness is open to discussion. Additionally, we will investigate the performance of SAITS on other real-world large datasets to further validate the model's generality in other domains.

\section*{Acknowledgments and Disclosure of Funding}
We sincerely appreciate Dr. Wei Cao at Microsoft Research Asia, Prof. Christian Desrosiers at \'ETS Montr\'eal, Dr. Ziyu Jia at Beijing Jiaotong University, and our anonymous reviewers for their constructive comments.
We thank Ciena for providing computing resources.

Wenjie Du is supported by a Mitacs accelerate program (\# FR61813) cooperating with Ciena.
This research is partially supported by the Natural Sciences and Engineering Research Council of Canada (\# RGPIN-2020-06797).

\clearpage

\bibliographystyle{unsrt}
\bibliography{references}

\clearpage
\appendix
\section{Details of Hyper-parameter Searching}\label{appendix: hyper-parameter search}
We expose the details about hyper-parameter searches in this section.

\textbf{General} \hspace{1em} 
For all models, the learning rate is log-uniformly sampled between $1 \times 10^{-4}$ and $1 \times 10^{-2}$. If applicable, the dropout rate is sampled from the values (0.0, 0.1, 0.2, 0.3, 0.4, 0.5).

\textbf{RNN-based models} \hspace{1em} 
For all RNN-based models (GRUI-GAN, E$^2$GAN, M-RNN, and BRITS), the RNN hidden size is sampled from the values (32, 64, 128, 256, 512, 1024). For GRUI-GAN and E$^2$GAN, the dimension of $z$ is sampled from (32, 64, 128, 256, 512, 1024), the number of pretrain epochs is sampled from (5, 10, 15, 20). For GRUI-GAN, hyper-parameter $\lambda$, which controls the proportion between the masked reconstruction loss and the discriminative loss, is sampled from (0, 0.15, 0.3, 0.45). For E$^2$GAN, hyper-parameter $\lambda$, which controls the weight of the discriminative loss and the squared error loss, is sampled from (2, 4, 8, 16, 32, 64).

\textbf{GP-VAE} \hspace{1em}
The encoder size and decoder size are sampled from values (64, 128, 256, 512, 1024). The length scale is sampled from (4, 8, 12, 16) and the window size is sampled from (4, 8, 16, 32, 64). $\beta$ is sampled from (0.1, 0.2, 0.3, 0.4, 0.5, 0.6, 0.7, 0.8). $\sigma$ is set as 1.005.

\textbf{Self-attention-based models} \hspace{1em} 
For self-attention models (Transformer and SAITS), we sample the number of layers $N$ from (1, 2, 3, 4, 5, 6, 7, 8), $d_\text{model}$ from (64, 128, 256, 512, 1024), $d_\text{ffn}$ from (128, 256, 512, 1024, 2048, 4096), $d_v$ from (32, 64, 128, 256, 512), the number of heads $h$ from (2, 4, 8). $d_k$ is set as the value of $d_\text{model}$ divided by $h$.

For model SAITS-base, we fix the learning rate = 0.001, the dropout rate = 0.1, $N = 2, d_\text{model} = 256, d_\text{ffn} = 128, h = 4, d_v = d_k = 64$.

\section{BRITS Trained by the joint-optimization approach} \label{BRITS_apply_MIT}
\begin{table} [ht]
	\caption{Performance comparison between BRITS trained without MIT and with MIT.}
	\label{BRITS_MIT_results}
	\centering
	\begin{minipage}{1\textwidth}
		\resizebox{163mm}{!}{
	\begin{tabular}{p{65pt}<{\centering}|p{90pt}<{\centering}|p{90pt}<{\centering}|p{90pt}<{\centering}|p{90pt}<{\centering}}
		\toprule
		Model         & PhysioNet-2012            & Air-Quality               & Electricity  & ETT \\
		\midrule
		BRITS-w/oMIT & 0.256 / 0.767 / 36.5\%    & 0.153 / 0.525 / 21.6\%    & \textbf{0.847 / 1.322 / 45.3\%}  & 0.130 / 0.259 / \textbf{12.5\%}  \\
		\midrule
		BRITS-wMIT  &\textbf{0.251 / 0.691 / 35.8\%}&\textbf{0.144 / 0.521 / 20.3\%}& 0.910 / 1.363 / 48.7\% & \textbf{0.123 / 0.237} / 12.6\% \\ 
		\bottomrule
	\end{tabular}
}
\end{minipage}
\end{table}

To discuss how our joint-optimization approach can influence the performance of RNN-based models, we apply it in the training of model BRITS and show experimental results in this section. 

BRITS models from Table~\ref{tb1} are used here. That is to say, hyper-parameters are kept exactly the same. The difference between the training way in the original paper~\cite{Cao2018BRITS} and our joint-optimization approach is whether to apply MIT. Consequently, we use suffix "w/oMIT" to represent the original training way and suffix "wMIT" to represent our joint-optimization training approach.

As displayed in Table~\ref{BRITS_MIT_results}, BRITS-wMIT outperforms BRITS-w/oMIT on datasets PhysioNet-2012, Air-Quality, and ETT, but achieves worse performance on the Electricity dataset. Therefore, applying MIT in the training of BRITS can bring further improvement on some datasets, but this is not necessary, and it depends on the dataset. Note that despite BRITS-wMIT obtains better results on datasets PhysioNet-2012, Air-Quality, and ETT, its performance is still inferior to Transformer and SAITS.	

\end{document}